\documentclass[english,preprint,12pt,authoryear]{elsarticle}

\usepackage[colorlinks=true]{hyperref}
\usepackage{geometry}
\usepackage{color}
\definecolor{revise_red}{rgb}{1,0,0} %

\usepackage{babel}
\usepackage{float}
\usepackage{graphicx}
\usepackage{etoolbox}
\usepackage{times}
\usepackage{epsfig}
\usepackage{graphicx}
\usepackage{url}
\usepackage{bibspacing}
\usepackage{amsfonts,bm,amsmath,mathtools,balance,algorithm,algorithmic,algorithm,amssymb,pgfplots,multirow,makecell,booktabs}

\usepackage{array}
\newcolumntype{I}{!{\vrule width 0.8pt}}

\journal{ISPRS Journal of Photogrammetry and Remote Sensing}

\begin{document}

\begin{frontmatter}

\renewcommand{\thefootnote}{\fnsymbol{footnote}}
\setcounter{footnote}{0}

\title{Semantic Labeling in Very High Resolution Images via A Self-Cascaded Convolutional Neural Network}
\author{Yongcheng\ Liu$^{a,b}$}
\author{Bin\ Fan$^{a,}$\footnote{Corresponding author at: National Laboratory of Pattern Recognition, Institute of Automation, Chinese Academy of Sciences, Beijing, P.R. China. E-mail address: bfan@nlpr.ia.ac.cn (B. Fan).}}
\author{Lingfeng\ Wang$^{a}$}
\author{Jun\ Bai$^{c}$}
\author{Shiming\ Xiang$^{a}$}
\author{Chunhong\ Pan$^{a}$}
\address{$^{a}$ National Laboratory of Pattern Recognition, Institute of Automation, Chinese Academy of Sciences, Beijing, P.R. China \\
$^{b}$ School of Computer and Control Engineering, University of Chinese Academy of Sciences, Beijing, P.R. China \\
$^{c}$ Research Center for Brain-inspired Intelligence, Institute of Automation, Chinese Academy of Sciences, Beijing, P.R. China}

\begin{abstract}
Semantic labeling for very high resolution (VHR) images in urban areas, is of significant importance in a wide range of remote sensing applications. However, many confusing manmade objects and intricate fine-structured objects make it very difficult to obtain both coherent and accurate labeling results. For this challenging task, we propose a novel deep model with convolutional neural networks (CNNs), i.e., an end-to-end self-cascaded network (ScasNet). Specifically, for confusing manmade objects, ScasNet improves the labeling coherence with sequential global-to-local contexts aggregation. Technically, multi-scale contexts are captured on the output of a CNN encoder, and then they are successively aggregated in a self-cascaded manner. Meanwhile, for fine-structured objects, ScasNet boosts the labeling accuracy with a coarse-to-fine refinement strategy. It progressively refines the target objects using the low-level features learned by CNN's shallow layers. In addition, to correct the latent fitting residual caused by multi-feature fusion inside ScasNet, a dedicated residual correction scheme is proposed. It greatly improves the effectiveness of ScasNet. Extensive experimental results on three public datasets, including two challenging benchmarks, show that ScasNet achieves the state-of-the-art performance.
\end{abstract}


\end{frontmatter}

\section{Introduction}
\label{Section1}
Semantic labeling in very high resolution (VHR) images is a long-standing research problem in remote sensing field. It plays a vital role in many important applications, such as infrastructure planning, territorial planning and urban change detection \citep{journals_toc_LuYZ2017,journals_rs_MatKK2011,journals_rse_ZhaSKC2011}. The target of this problem is to assign each pixel to a given object category. Note that it is not just limited to building extraction \citep{journals_tgrs_LiFXZW2015}, road extraction \citep{journals_tgrs_chengWXWXP2017} and vegetation extraction \citep{journals_ijeors_WenLLZ2017} which only consider labeling one single category, semantic labeling usually considers several categories simultaneously \citep{journals_tgrs_LiHGDZBP2015,journals_tgrs_XuLHMP2016,journals_tgrs_XueLCD2015}. As a result, this task is very challenging, especially for the urban areas, which exhibit high diversity of manmade objects. Specifically, on one hand, many manmade objects (e.g., buildings) show various structures, and they are composed of a large number of different materials. Meanwhile, plenty of different manmade objects (e.g., buildings and roads) present much similar visual characteristics. These confusing manmade objects with high intra-class variance and low inter-class variance bring much difficulty for coherent labeling. On the other hand, fine-structured objects in cities (e.g., cars, trees and low vegetations) are quite small or threadlike, and they also interact with each other through occlusions and cast shadows. These factors always lead to inaccurate labeling results. Furthermore, it poses additional challenge to simultaneously label all these size-varied objects well.

To accomplish such a challenging task, features at different levels are required. Specifically, abstract high-level features are
more suitable for the recognition of confusing manmade objects, while labeling of fine-structured objects could benefit from detailed low-level features. Convolutional neural networks (CNNs) \citep{conf_lecunBDHHHJ1990} in \emph{deep learning} field are well-known for feature learning \citep{journals_ijrs_masJFJ2008}. CNNs consist of multiple trainable layers which can extract expressive features of different levels \citep{conf_lecunBBH_1998}. Moreover, recently, CNNs with \emph{deep learning} have demonstrated remarkable learning ability in computer vision field, such as scene recognition \citep{journals_tnnls_YuanML2015} and image segmentation \citep{conf_cvpr_LongSD15}. Meanwhile, the development of remote sensing has also been greatly promoted by numerous CNNs-based methods \citep{journals_corr_GongHL2017}. For example, deconvolution networks \citep{conf_cvpr_ZeilerKTF2010} are investigated by Lu et al. \citep{journals_tgrs_LuZY2017} for remote sensing scene classification, and Chen et al. \citep{journals_tgrs_ChenWXJ16} perform target classification using CNNs for SAR Images.

Based on CNNs, many patch-classification methods are proposed to perform semantic labeling \citep{thesis_mnih2013,conf_cvpr_mostaMYPSG2015,journals_ijeors_paiSSHA2016,conf_icpr_nogueMJWJ2016,journals_jprs_AlsPWM2017,journals_jprs_zhangXHAIJP2017}. These methods determine a pixel's label by using CNNs to classify a small patch around the target pixel. However, they are far from optimal, because they ignore the inherent relationship between patches and their time consumption is huge \citep{journals_tgrs_magETYCGAP2017}. Typically, fully convolutional networks (FCNs) have boosted the accuracy of semantic labeling a lot \citep{conf_cvpr_LongSD15,journals_corr_sheJ2016}. FCNs perform pixel-level classification directly and now become the normal framework for semantic labeling. Nevertheless, due to multiple \emph{sub-samplings} in FCNs, the final \emph{feature maps} are much coarser than the input image, resulting in less accurate labeling results.

Accordingly, a tough problem locates on how to perform accurate labeling with the coarse output of FCNs-based methods, especially for fine-structured objects in VHR images. To solve this problem, some researches try to reuse the low-level features learned by CNNs' shallow layers \citep{conf_eccv_ZeilerMDFR2014}. The aim is to utilize the local details (e.g., corners and edges) captured by the \emph{feature maps} in fine resolution. Technically, they perform operations of multi-level feature fusion \citep{conf_miccai_RonnebergerFB15,conf_cvpr_LongSD15,conf_cvpr_harihaAPAGRMJ2015,journals_corr_PinPLTCRD2016}, \emph{deconvolution} \citep{conf_iccv_NohHH15} or \emph{up-pooling} with recorded \emph{pooling} indices \citep{journals_corr_BadrinarayananK15}. Most of these methods use the strategy of direct stack-fusion. However, this strategy ignores the inherent semantic gaps in features of different levels. An alternative way is to impose boundary detection \citep{conf_cvpr_BerGSJL2016,journals_corr_MarSWJSU2016}. It usually requires extra boundary supervision and leads to extra model complexity despite boosting the accuracy of object localization.

Another tricky problem is the labeling incoherence of confusing objects, especially of the various manmade objects in VHR images. To tackle this problem, some researches concentrate on leveraging the multi-context to improve the recognition ability of those objects. They use multi-scale images \citep{journals_pami_FarabCNLY2013,conf_cvpr_mostaMYPSG2015,journals_nc_chengZXWP2016,conf_igarss_LiuYFZ2016,conf_cvpr_chenYXY2016,journals_jprs_zhaoZDS2016} or multi-region images \citep{conf_iccv_GidarKN2015,journals_luusSFB2015} as input to CNNs. However, these methods are usually less efficient due to a lot of repetitive computation. Differently, some other researches are devoted to acquire multi-context from the inside of CNNs. They usually perform operations of multi-scale \emph{dilated convolution} \citep{conf_iclr_ChenPKMY15}, multi-scale pooling \citep{journal_kaimingZRS2015,conf_iclr_LiuRABA2016,conf_cvpr_bellCBKGR2016} or multi-kernel convolution \citep{journals_corr_AudebSLLS2016}, and then fuse the acquired multi-scale contexts in a direct stack manner. Nevertheless, this manner not only ignores the hierarchical dependencies among the objects and scenes in different scales, but also neglects the inherent semantic gaps in contexts of different-level information.

In summary, although current CNN-based methods have achieved significant breakthroughs in semantic labeling, it is still difficult to label the VHR images in urban areas. The reasons are as follows: 1) Most existing approaches are less efficient to acquire multi-scale contexts for confusing manmade objects recognition; 2) Most existing strategies are less effective to utilize low-level features for accurate labeling, especially for fine-structured objects; 3) Simultaneously fixing the above two issues with a single network is particularly difficult due to a lot of fitting residual in the network, which is caused by semantic gaps in different-level contexts and features.

\begin{figure*}[t]
\centerline{\includegraphics[width=18.15cm]{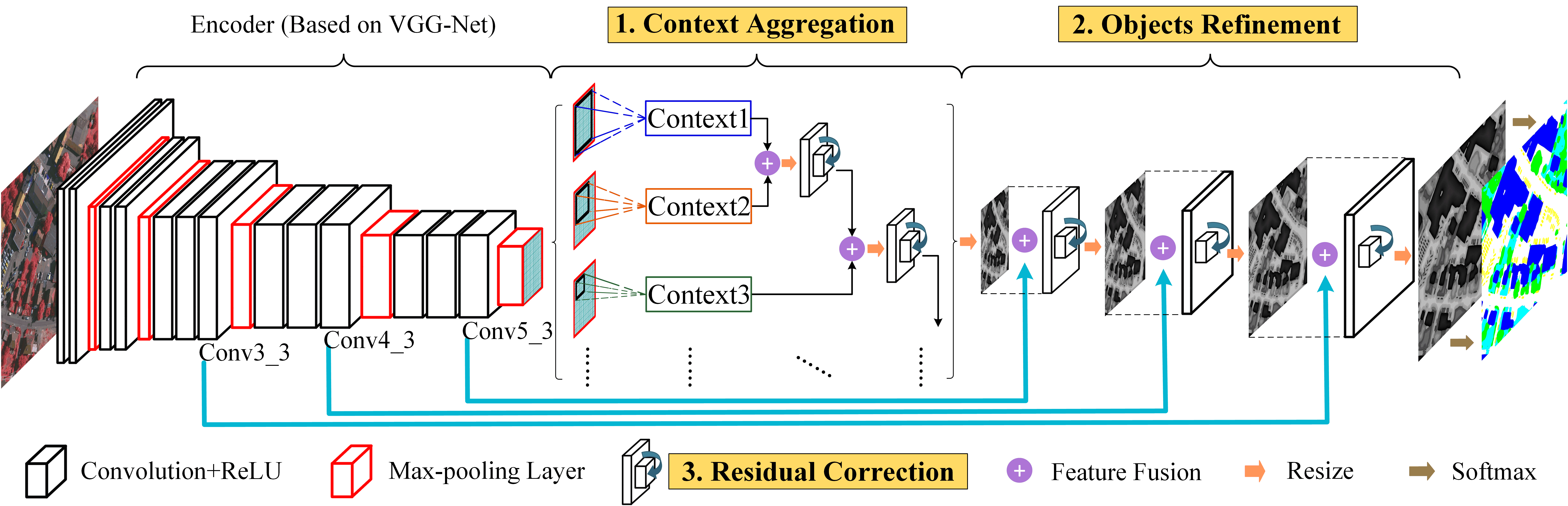}}
\caption{Overview of the proposed ScasNet. (Best viewed in color)}
\label{fig:scasnet_overview}
\end{figure*}

In this paper, we propose a novel self-cascaded convolutional neural network (ScasNet), as illustrated in Fig. \ref{fig:scasnet_overview}. The aim of this work is to further advance the state of the art on semantic labeling in VHR images. To this end, it is focused on three aspects: 1) multi-scale contexts aggregation for distinguishing confusing manmade objects; 2) utilization of low-level features for fine-structured objects refinement; 3) residual correction for more effective multi-feature fusion. Specifically, a conventional CNN is adopted as an encoder to extract features of different levels. On the \emph{feature maps} outputted by the encoder, global-to-local contexts are sequentially aggregated for confusing manmade objects recognition. Technically, multi-scale contexts are first captured by different convolutional operations, and then they are successively aggregated in a self-cascaded manner. With the acquired contextual information, a coarse-to-fine refinement strategy is performed to refine the fine-structured objects. It progressively reutilizes the low-level features learned by CNN's shallow layers with long-span connections. In addition, to correct the latent fitting residual caused by semantic gaps in multi-feature fusion, several residual correction schemes are employed throughout the network. As a result of residual correction, the above two different solutions could work collaboratively and effectively when they are integrated into a single network. Extensive experiments demonstrate the effectiveness of ScasNet. Moreover, the three submodules in ScasNet could not only provide good solutions for semantic labeling, but are also suitable for other tasks such as object detection \citep{journals_jprs_GongJ2016} and change detection \citep{journals_jprs_zhangMLJZ2016,journals_jprs_GongHP2017}, which will no doubt benefit the development of the remote sensing deep learning techniques.

To sum up, the main contributions of this paper can be highlighted as follows:
\begin{itemize}
\setlength{\itemsep}{0ex}
\item A self-cascaded architecture is proposed to successively aggregate contexts from large scale to small ones. In this way, global-to-local contexts with hierarchical dependencies among the objects and scenes are well retained, resulting in coherent labeling results of confusing manmade objects.
\item A coarse-to-fine refinement strategy is proposed, which progressively refines the target objects using the low-level features learned by CNN's shallow layers. Thus, accurate labeling results can be achieved, especially for the fine-structured objects.
\item A residual correction scheme is proposed to correct the latent fitting residual caused by semantic gaps in multi-feature fusion. It greatly improves the effectiveness of the above two different solutions.
\item All the above contributions constitute a novel end-to-end \emph{deep learning} framework for semantic labelling, as shown in Fig. \ref{fig:scasnet_overview}. It achieves the state-of-the-art performance on two challenging benchmarks by the date of submission: \emph{ISPRS 2D Semantic Labeling Challenge} \citep{ISPRS_challenge} for Vaihingen and Potsdam. Furthermore, these results are obtained using only image data with a single model, without using the elevation data like the Digital Surface Model (DSM), model ensemble strategy or any postprocessing.
\end{itemize}

A shorter version of this paper appears in \citep{conf_icip_liuFWXP2017}. Apart from extensive qualitative and quantitative evaluations on the original dataset, the main extensions in the current work are:
\begin{itemize}
\setlength{\itemsep}{0ex}
\item More comprehensive and elaborate descriptions about the proposed semantic labeling method.
\item Further performance improvement by the modification of network structure in ScasNet.
\item Comparative experiments with more state-of-the-art methods on another two challenging datasets for further support the effectiveness of ScasNet.
\item More detailed and in-depth analyses, as well as model visualization and complexity analyses of ScasNet.
\end{itemize}

The remainder of this paper is arranged as follows. The basic modules used in ScasNet are briefly introduced in Section \ref{Section2}. Section \ref{Section3} presents the details of the proposed semantic labeling method. Experimental evaluations between our method and the state-of-the-art methods, as well as detailed analyses of ScasNet are provided in Section \ref{Section4}. Finally, the conclusion is outlined in Section \ref{Section5}.

\section{Preliminaries}
\label{Section2}

CNNs \citep{conf_lecunBDHHHJ1990} are multilayer neural networks that can hierarchically extract powerful low-level and high-level features. The input and output of each layer are sets of arrays called \emph{feature maps}. Commonly, a standard CNN contains three kinds of layers: convolutional layer, nonlinear layer and pooling layer. The convolutional layer offers filter-like function to generate convoluted \emph{feature maps}, while the nonlinear layer simply consists of an elementwise nonlinear activation function applied to each value in the \emph{feature maps}. The pooling layer generalizes the convoluted features into higher level, which makes features more abstract and robust. Meanwhile, in CNNs, the feature extraction module and the classifier module are integrated into one framework, thus the extracted features are more suitable for specific task than hand-crafted features, such as HOG \citep{conf_cvpr_DalalT05}, SIFT \citep{journals_ijcv_Lowe04}, and spectral features in remote sensing \citep{journals_tgrs_zhangZTH2012}.

In the following, each basic layer used in the proposed network will be introduced, and their specific configurations will be presented in Section \ref{subsection:scasnet_config}.

\begin{enumerate}[]
\item {\bf Convolutional Layer:} The convolutional (Conv) layer performs a series of convolutional operations on the previous layer with a small kernel (e.g., $3\times3$). The output of each convolutional operation is computed by dot product between the weights of the kernel and the corresponding local area (\emph{local receptive field}). A \emph{weight sharing} technique that the parameters (i.e., weights and bias) are shared among each kernel across an entire \emph{feature map}, is adopted to reduce parameters in great deal \citep{journals_Nature_RumelhartHW1986}.

\item {\bf Batch Normalization Layer:} Batch normalization (BN) mechanism \citep{conf_icml_IoffeS15} normalizes layer inputs to a Gaussian distribution with zero-mean and unit variance, aiming at addressing the problem of \emph{internal covariate shift}, i.e., the distribution of each layer's inputs changes during training, as the parameters of the previous layers change. Thus, it allows us to use much higher learning rate.

\item {\bf ReLU Layer:} The rectified linear unit (ReLU) \citep{journals_aista_GloXBBY2011,conf_icml_NairVHE2010} is usually chosen as the nonlinearity layer. It thresholds the non-positive value as zero and keeps the positive value unchanged, i.e., an elementwise activation as $\max(0,x)$. ReLU can achieve a considerable reduction in training time \citep{conf_nips_KrizhevskySH12}.

\item {\bf Pooling Layer:} Pooling is a way to perform \emph{sub-sampling} along the spatial dimension. Commonly, there are two kinds of pooling: max-pooling and ave-pooling. Max-pooling samples the maximum in the region to be pooled, while ave-pooling computes the mean value. In our network, we use max-pooling.

\item {\bf Dropout Layer:} Dropout \citep{journals_jmlr_Sriva2014} is an effective regularization technique to reduce overfitting. It randomly drops units (along with their connections) from the neural network during training, which prevents units from co-adapting too much.

\item {\bf Interpolation Layer:} Interpolation (Interp) layer performs resizing operation along the spatial dimension. In our network, we use bilinear interpolation.

\item {\bf Elementwise Layer:} Elementwise (Eltwise) layer performs elementwise operations on two or more previous layers, in which the \emph{feature maps} must be of the same number of channels and the same size. There are three kinds of elementwise operations: \emph{product, sum, max}. In our network, we use \emph{sum} operation.

\item {\bf Softmax Layer:} The softmax nonlinearity \citep{journals_softmax_1989} is applied to the output layer in
the case of multiclass classification. It outputs the posterior probabilities over each category.
\end{enumerate}

\section{Self-cascaded Convolutional Neural Network (ScasNet)}
\label{Section3}

Semantic labeling also called pixel-level classification, is aimed at obtaining all the pixel-level categories in an entire image.
For this task, we have to predict the most likely category $\hat{k}$ for a given image $x$ at $j$-th pixel $x^j$, which is given by
\begin{equation}
\label{Eq:problem_formula}
\hat{k} = \underset{k\in \mathcal{C}}{\rm{argmax}}\;p_k(x^j|{\bm{\theta}}), \; \forall \, j\in\{1,\cdots,N\},
\end{equation}
where $p_k(x^j|{\bm{\theta}})$, estimated by a model with parameters $\bm{\theta}$, denotes the posterior probability of $x^j$ belonging to the $k$-th category in a set of categories $\mathcal{C}=\{1,\cdots,K\}$. $K$ is the number of categories and $N$ is the number of pixels in the given image.

In this work, we perform semantic labeling for VHR images in urban areas by means of a self-cascaded convolutional neural network (ScasNet), which is illustrated in Fig. \ref{fig:scasnet_overview}. In the following, we will describe five important aspects of ScasNet, including 1) \emph{Multi-scale contexts Aggregation}, 2) \emph{Fine-structured Objects Refinement}, 3) \emph{Residual Correction}, 4) \emph{ScasNet Configuration}, 5) \emph{Learning and Inference Algorithm}.

\subsection{\bf Multi-scale contexts Aggregation}

Obtaining coherent labeling results for confusing manmade objects in VHR images is not easily accessible, because they are of high intra-class variance and low inter-class variance. To fix this issue, it is insufficient to use only the very local information of the target objects. We need to know the scene information around them, which could provide much wider visual cues to better distinguish the confusing objects. The scene information also means the context, which characterizes the underlying dependencies between an object and its surroundings, is a critical indicator for objects identification. Therefore, we are interested in discussing how to efficiently acquire context with CNNs in this Section.

In CNNs, each unit of deeper layers (\emph{feature maps}) contains more extensive, powerful and abstract information, due to the larger \emph{receptive field} on the input image and higher nonlinearity \citep{conf_eccv_ZeilerMDFR2014}. Thus, the context acquired from deeper layers can capture wider visual cues and stronger semantics simultaneously. However, only single-scale context may not represent hierarchical dependencies between an object and its surroundings. Naturally, multi-scale contexts are gaining more attention. However, it is very hard to retain the hierarchical dependencies in contexts of different scales using common fusion strategies (e.g., direct stack). To address this issue, we propose a novel self-cascaded architecture, as shown in the middle part of Fig. \ref{fig:scasnet_overview}. It is aimed at aggregating global-to-local contexts while well retaining hierarchical dependencies, i.e., the underlying inclusion and location relationship among the objects and scenes in different scales (e.g., the car is more likely on the road, the chimney and skylight is more likely a part of roof and the roof is more likely by the road).

\begin{figure}[t]
\begin{minipage}[b]{0.41\linewidth}
  \centering
  \centerline{\includegraphics[width=7.45cm]{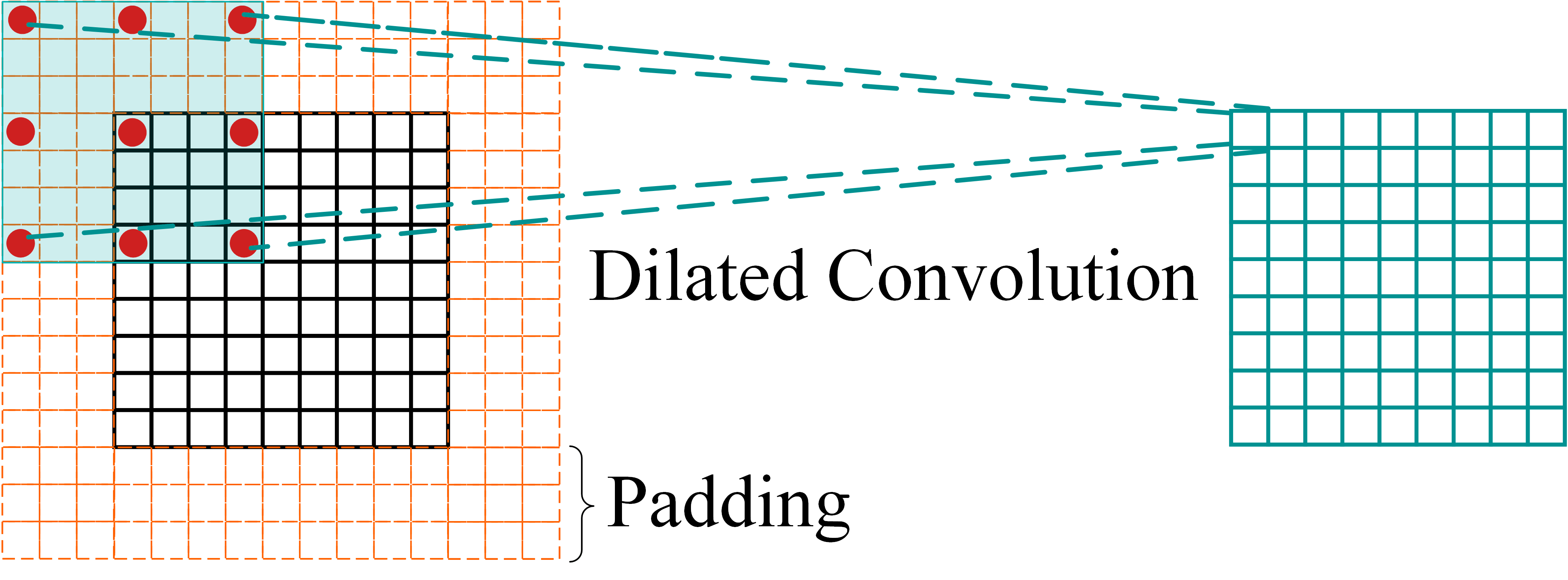}}
  \centerline{(a)}
\end{minipage}
\begin{minipage}[b]{0.33\linewidth}
  \centering
  \centerline{\includegraphics[width=6.1cm]{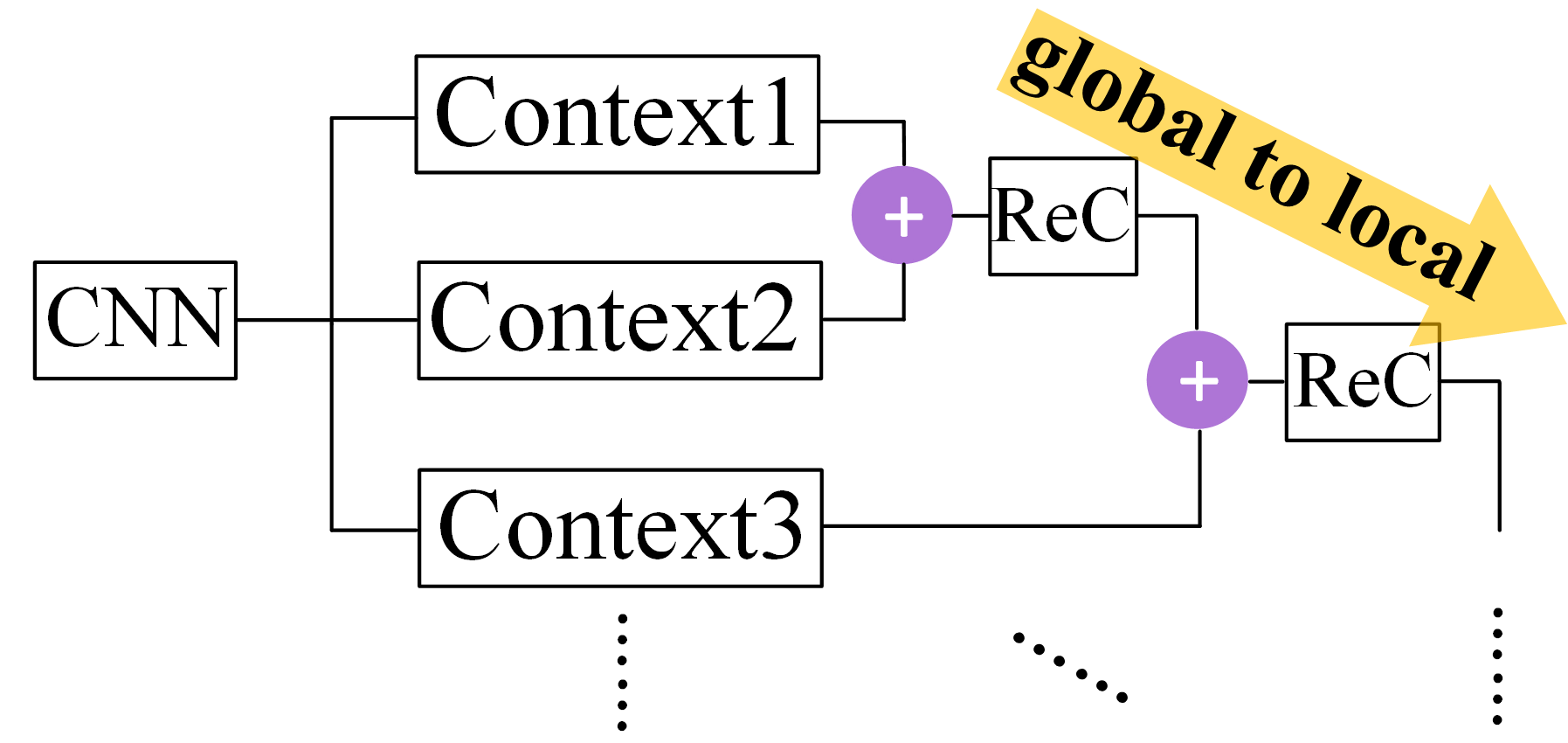}}
  \centerline{(b)}
\end{minipage}
\begin{minipage}[b]{0.25\linewidth}
  \centering
  \centerline{\includegraphics[width=4.2cm]{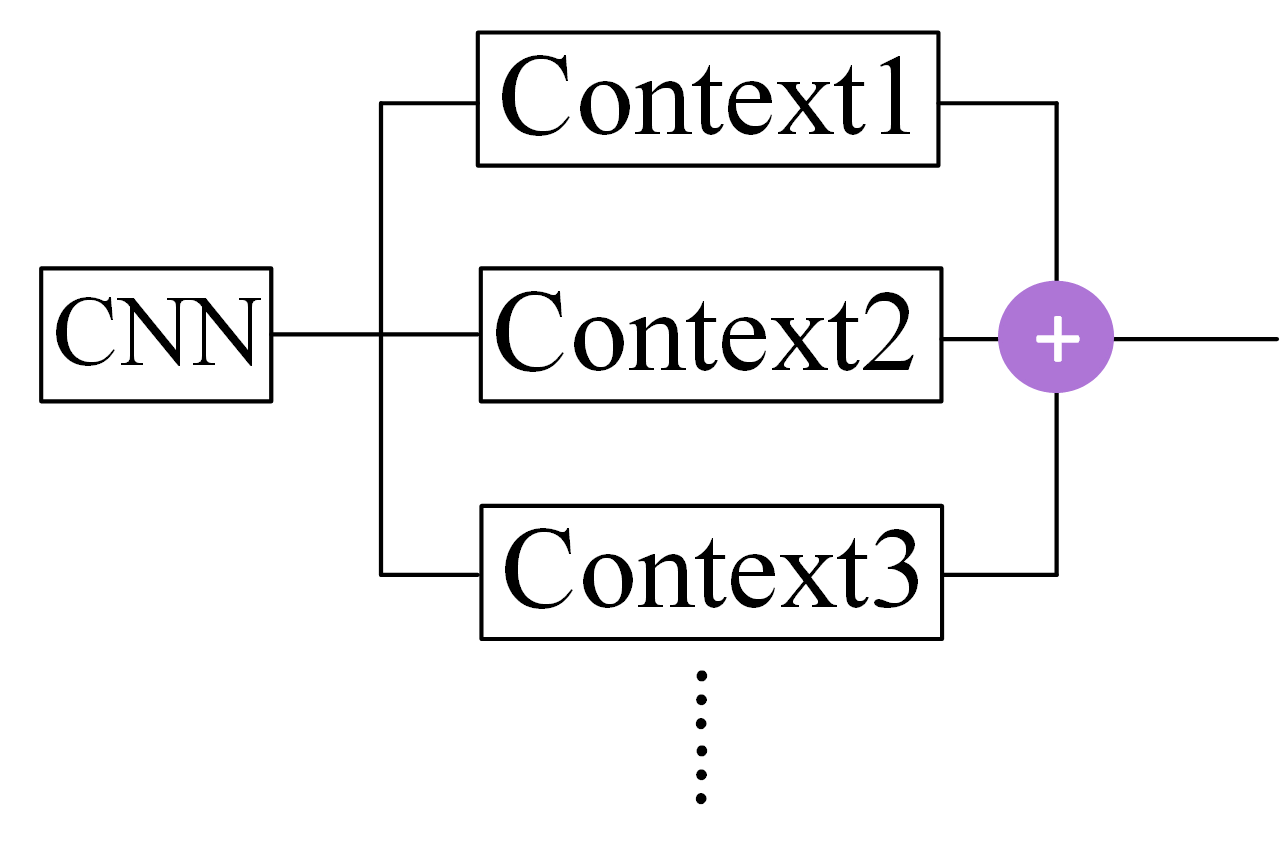}}
  \centerline{(c)}
\end{minipage}

\caption{(a) An illustration of \emph{dilated convolution} used in ScasNet to capture context, where the size of \emph{feature map} and \emph{convolution kernel} is $9\times9$ and $3\times3$, respectively, both the \emph{dilation rate} and the \emph{padding rate} equal $3$, and the \emph{padding value} is zero. (b) The proposed multi-context aggregation approach, i.e., performing aggregation sequentially in a self-cascaded manner. (c) Multi-context aggregation in a parallel stack. `ReC' denotes the proposed residual correction scheme. (Best viewed in color)}
\label{fig:context}
\end{figure}

Specifically, we perform \emph{dilated convolution} operation on the last layer of the encoder to capture context. The reasons are two-fold. On one hand, dilated convolution expands the \emph{receptive field}, which can capture high-level semantics with wider information. On the other hand, although theoretically, features from high-level layers of a network have very large \emph{receptive fields} on the input image, in practice they are much smaller \citep{conf_zhou_2015ICLR}. This problem can be alleviated by \emph{dilated convolution}. Fig. \ref{fig:context}(a) illustrates an example of \emph{dilated convolution}. To make the size of \emph{feature map} after \emph{dilated convolution} unchanged, the \emph{padding rate} should be set as the same to the \emph{dilation rate}. More details about \emph{dilated convolution} can be referred in \citep{conf_iclr_YuFKV2016}.

\renewcommand{\thefootnote}{\arabic{footnote}}
\setcounter{footnote}{0}
Then, by setting a group of big-to-small \emph{dilation rates} (24, 18, 12 and 6 in the experiment), a series of \emph{feature maps} with global-to-local contexts are generated \footnote{Due to the inherent properties of convolutional operation in each single-scale context (same-scale convolution kernels with large original receptive fields convolve with weight sharing over spatial dimension and summation over channel dimension), the relationship between contexts with same scale can be acquired implicitly.}. That is, multi-scale \emph{dilated convolution} operations correspond to multi-size regions on the last layer of encoder (see Fig. \ref{fig:scasnet_overview}). Large region (high-level context) contains more semantics and wider visual cues, while small region (low-level context) otherwise. Meanwhile, the obtained \emph{feature maps} with multi-scale contexts can be aligned automatically due to their equal resolution.

To well retain the hierarchical dependencies in multi-scale contexts, we sequentially aggregate them from global to local in a self-cascaded manner as shown in Fig. \ref{fig:context}(b). In this way, high-level context with big \emph{dilation rate} is aggregated first and low-level context with small \emph{dilation rate} next. Formally, it can be described as:
\begin{equation}\label{Eq:context}
\left\{ \begin{array}{l}
  \mathrm{T} = \Upsilon \bigg[ \cdots \Upsilon \Big[ \Upsilon [\mathrm{T}_1 \oplus \mathrm{T}_2] \oplus \mathrm{T}_3 \Big] \oplus \cdots \oplus \mathrm{T}_n \bigg], \\
  d_{_{\mathrm{T}_1}} > d_{_{\mathrm{T}_2}} > d_{_{\mathrm{T}_3}} > \cdots > d_{_{\mathrm{T}_n}}.
\end{array} \right.
\end{equation}
Here, $\mathrm{T}_1, \mathrm{T}_2, \cdots, \mathrm{T}_n$ denote $n$-level contexts, $\mathrm{T}$ is the final aggregated context and $d_{T_i}$ ($i=1,\dots,n$) is the \emph{dilation rate} set for capturing the context $\mathrm{T}_i$. `$\oplus$' denotes the fusion operation. $\Upsilon[\cdot]$ denotes the residual correction process, which will be described in Section \ref{subsection:residual_correct}. In fact, the above aggregation rule is consistent with the visual mechanism, i.e., wider visual cues in high-level context could play a guiding role in integrating low-level context. For instance, the visual impression of a whole roof can provide strong guidance for the recognition of chimney and skylight in this roof.

The proposed self-cascaded architecture for multi-scale contexts aggregation has several advantages: 1) The multiple contexts are acquired from deep layers in CNNs, which is more efficient than directly using multiple images as input \citep{conf_iccv_GidarKN2015}; 2) Besides the hierarchical visual cues, the acquired contexts also capture the abstract semantics learned by CNN, which is more powerful for confusing objects recognition; 3) The self-cascaded strategy of sequentially aggregating multi-scale contexts, is more effective than the parallel stacking strategy \citep{conf_iclr_ChenPKMY15,conf_iclr_LiuRABA2016}, as shown in Fig. \ref{fig:context}(c), which potentially loses the hierarchical dependencies in different scales; 4) The more complicated nonlinear operation of Eq. \eqref{Eq:context} has a stronger capacity to fit the underlying mapping than those stacking operations.

\subsection{\bf Fine-structured Objects Refinement}

Besides the complex manmade objects, intricate fine-structured objects also increase the difficulty for accurate labeling in VHR images. Actually, the final \emph{feature maps} outputted by the FCN-based methods is quite coarse due to multiple \emph{sub-samplings}. For example, the size of the last \emph{feature maps} in VGG-Net \citep{journals_corr_SimonyanZ14a} is $1/32$ of input size. Thus, it is very hard to restore the low-level details of objects (e.g., boundary and localization) for accurate labeling, especially for fine-structured objects.

\begin{figure}[t]
\begin{minipage}[b]{1\linewidth}
  \centering
  \centerline{\includegraphics[width=11cm]{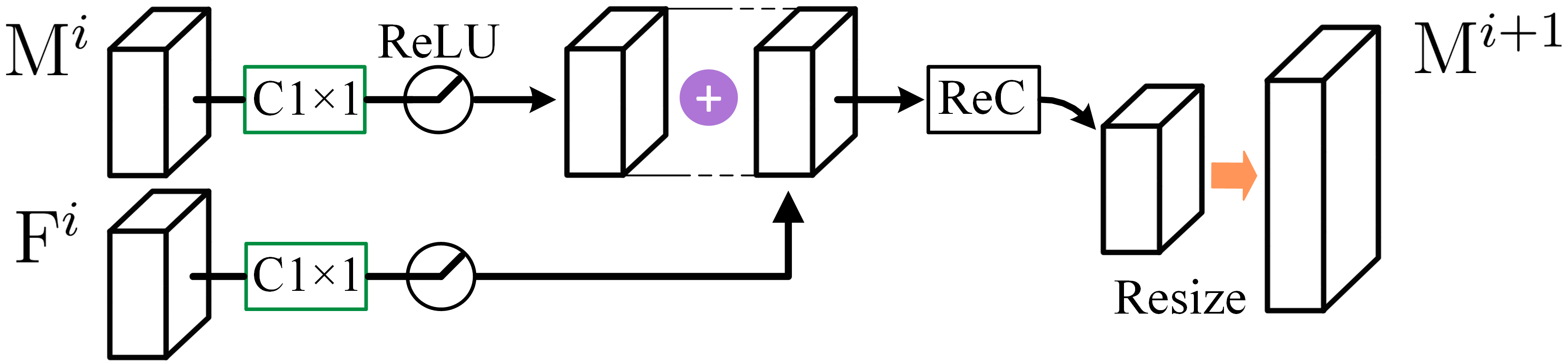}}
\end{minipage}

\caption{Single process of refinement. `C1$\times$1' denotes convolutional operation with \emph{kernel size} 1$\times$1, `ReC' denotes the residual correction scheme. (Best viewed in color)}
\label{fig:refine}
\end{figure}

In CNNs, it is found that the low-level features can usually be captured by the shallow layers \citep{conf_eccv_ZeilerMDFR2014}. Based on this observation, we propose to reutilize the low-level features with a coarse-to-fine refinement strategy, as shown in the rightmost part of Fig. \ref{fig:scasnet_overview}. Specifically, the shallow layers with fine resolution are progressively reintroduced into the decoder stream by long-span connections. As a result, the coarse \emph{feature maps} can be refined and the low-level details can be recovered. Each single refinement process is illustrated in Fig. \ref{fig:refine}, which can be formulated as:
\begin{equation}\label{Eq:refine}
  \mathrm{M}^{i+1} =  \Re\bigg[ \Upsilon \Big[ \mathcal{L}(\mathrm{M}^{i} \otimes \mathbf{w}_{\mathrm{M}^{i}}) \oplus \mathcal{L}(\mathrm{F}^{i} \otimes \mathbf{w}_{\mathrm{F}^{i}}) \Big] \bigg],
\end{equation}
where $\mathrm{M}^{i}$ denotes the refined feature maps of the previous process, and $\mathrm{F}^{i}$ denotes the feature maps to be reutilized in this process coming from a shallower layer. $\mathbf{w}_{\mathrm{M}^{i}}$ and $\mathbf{w}_{\mathrm{F}^{i}}$ are the convolutional weights for $\mathrm{M}^{i}$ and $\mathrm{F}^{i}$ respectively. `$\otimes$' and `$\oplus$' denote the operations of convolution and fusion, respectively. $\mathcal{L}(\cdot)$ is the ReLU activation function. $\Re[\cdot]$ denotes the resize process and $\Upsilon[\cdot]$ denotes the process of residual correction. To fuse finer detail information from the next shallower layer, we resize the current feature maps to the corresponding higher resolution with bilinear interpolation to generate $\mathrm{M}^{i+1}$.

It is fairly beneficial to fuse those low-level features using the proposed refinement strategy. On one hand, in fact, the \emph{feature maps} of different resolutions in the encoder (see Fig. \ref{fig:scasnet_overview}) represent semantics of different levels \citep{conf_eccv_ZeilerMDFR2014}. Thus, due to their inherent semantic gaps, stacking all these features directly \citep{conf_cvpr_harihaAPAGRMJ2015,journals_pami_FarabCNLY2013} may not be a good choice. In our method, the influence of semantic gaps is alleviated when a gradual fusion strategy is used. On the other hand, in training stage, the long-span connections allow direct gradient propagation to shallow layers, which helps effective end-to-end training.

The most relevant work with our refinement strategy is proposed in \citep{journals_corr_PinPLTCRD2016}, however, it is different from ours to a large extent. On one hand, our strategy focuses on performing dedicated refinement considering the specific properties (e.g., small dataset and intricate scenes) of VHR images in urban areas. Specifically, as shown in Fig. \ref{fig:scasnet_overview}, only a few specific shallow layers are chosen for the refinement. Those layers that actually contain adverse noise due to intricate scenes are not incorporated. On the other hand, our refinement strategy works with our specially designed residual correction scheme, which will be elaborated in the following Section.

\subsection{\bf Residual Correction}
\label{subsection:residual_correct}
\begin{figure}[t]
\begin{minipage}[t]{1\linewidth}
  \centering
  \centerline{\includegraphics[width=10cm]{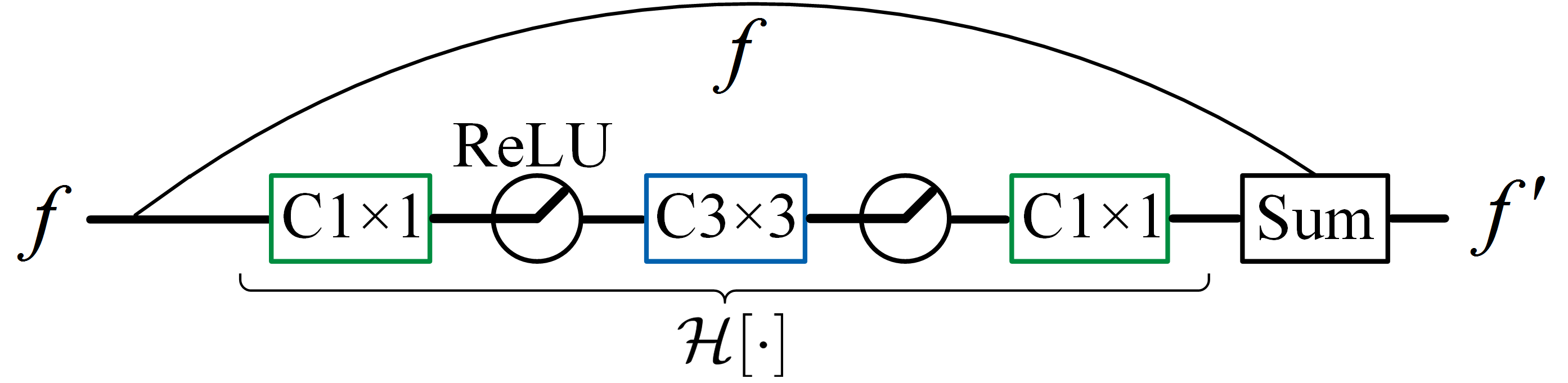}}
\end{minipage}

\caption{Residual correction scheme. `C1$\times$1' and `C3$\times$3' denote convolutional operation with \emph{kernel size} 1$\times$1 and 3$\times$3, respectively. (Best viewed in color)}
\label{fig:residual}
\end{figure}

It is notable that the proposed two solutions for labeling confusing manmade objects and fine-structured objects are quite different. In order to collaboratively and effectively integrate them into a single network, we have to find a approach to perform effective multi-feature fusion inside the network. This task is very challenging due to two issues. Firstly, as network deepens, it is fairly difficult for CNNs to directly fit a desired underlying mapping \citep{conf_cvpr_HeZRS2016}. Furthermore, this problem is worsened when it comes to fuse features of different levels. Secondly, there exists latent fitting residual when fusing multiple features of different semantics, which could cause the lack of information in the progress of fusion. To address this problem, a residual correction scheme is proposed, as shown in Fig. \ref{fig:residual}. It is dedicatedly aimed at correcting the latent fitting residual in multi-feature fusion inside ScasNet.

Specifically, building on the idea of deep residual learning \citep{conf_cvpr_HeZRS2016}, we explicitly let the stacked layers fit an inverse residual mapping, instead of directly fitting a desired underlying fusion mapping. Formally, let $f$ denote fused feature and $f^{'}$ denote the desired underlying fusion. We expect the stacked layers to fit another mapping, which we call inverse residual mapping as:
\begin{equation}\label{Eq:residual}
  \mathcal{H}[\cdot] = f^{'} - f.
\end{equation}

Actually, the aim of $\mathcal{H}[\cdot]$ is to compensate for the lack of information caused by the latent fitting residual, thus to achieve the desired underlying fusion $f^{'}= f + \mathcal{H}[\cdot]$. Moreover, as demonstrated by \citep{conf_cvpr_HeZRS2016}, the inverse residual learning can be very effective in deep network, because it is easier to fit $\mathcal{H}[\cdot]$ than to directly fit $f^{'}$ when network deepens. As a result, the adverse influence of latent fitting residual in multi-feature fusion can be well counteracted, i.e, the residual is well corrected.

It should be noted that, our residual correction scheme is quite different from the so-called chained residual pooling in RefineNet \citep{journals_corr_LinMSR2016} on both function and structure. Functionally, the chained residual pooling in RefineNet aims to capture background context. However, our scheme explicitly focuses on correcting the latent fitting residual, which is caused by semantic gaps in multi-feature fusion. Structurally, the chained residual pooling is fairly complex, while our scheme is
simple and efficient. As can be seen in Fig. \ref{fig:residual}, only one basic residual block is used in our scheme, and it is simply constituted by three convolutional layers and a skip connection.

As shown in Fig. \ref{fig:scasnet_overview}, several residual correction modules are elaborately embedded in ScasNet, which can
greatly prevent the fitting residual from accumulating. As a result, the proposed two different solutions work collaboratively and effectively, leading to a very valid global-to-local and coarse-to-fine labeling manner. Besides, the skip connection (see Fig. \ref{fig:residual}) is very beneficial for gradient propagation, resulting in an efficient end-to-end training of ScasNet.

\subsection{\bf ScasNet Configuration}
\label{subsection:scasnet_config}
As depicted in Fig. \ref{fig:scasnet_overview}, the encoder network corresponds to a feature extractor that transforms the input image to multi-dimensional shrinking \emph{feature maps}. To achieve this function, any existing CNN structures can be taken as the encoder part. In this paper, we propose two types of ScasNet based on two typical networks, i.e., $16$-layer VGG-Net \citep{journals_corr_SimonyanZ14a} and $101$-layer ResNet \citep{conf_cvpr_HeZRS2016}. Compared with VGG ScasNet, ResNet ScasNet has better performance while suffering higher complexity.

\renewcommand{\thefootnote}{\fnsymbol{footnote}}
\setcounter{footnote}{0}
We supply the trained models of these two CNNs so that the community can directly choose one of them based on different applications which require different trade-off between accuracy and complexity. All codes of the two specific ScasNet are released on the github\footnote{\url{https://github.com/Yochengliu/ScasNet}}. For clarity, we briefly introduce their configurations in the following.

{\bf VGG ScasNet:} In VGG ScasNet, the encoder is based on a VGG-Net variant \citep{conf_iclr_ChenPKMY15}, which is to obtain finer \emph{feature maps} (about $1/8$ of input size rather than $1/32$). On the last layer of encoder, multi-scale contexts are captured by \emph{dilated convolution} operations with \emph{dilation rates} of $24$, $18$, $12$ and $6$. We only choose three shallow layers for refinement as shown in Fig. \ref{fig:scasnet_overview}. There are two reasons: 1) shallower layers also carry much adverse noise despite of finer low-level details contained in them; 2) It is very difficult to train a more complex network well with remote sensing datasets, which are usually very small. In the encoder, we always use the last convolutional layer in each stage prior to pooling for refinement, because they contain stronger semantics in that stage. Six residual correction modules are employed for multi-feature fusion. Finally, a softmax classifier is employed to obtain probability maps, which indicate the likelihood of each pixel belonging to a category.

{\bf ResNet ScasNet:} The configuration of ResNet ScasNet is almost the same as VGG ScasNet, except for four aspects: the encoder is based on a ResNet variant \citep{journals_corr_ZhaoSQWJ2016}, four shallow layers are used for refinement, seven residual correction modules are employed for feature fusions and BN layer is used.
{\renewcommand\baselinestretch{1}\selectfont
\renewcommand{\algorithmicrequire}{\textbf{Input:}}
\renewcommand{\algorithmicensure}{\textbf{Output:}}
\renewcommand{\algorithmicrepeat}{\textbf{Repeat:}}
\renewcommand{\algorithmicuntil}{\textbf{Until:}}
\renewcommand{\algorithmicreturn}{\textbf{Return:}}
\begin{algorithm}[t]
\caption{\ Learning procedure of the proposed ScasNet}
\label{alg:learning}
\small
\begin{algorithmic}[1]
\REQUIRE The image and label data $(\textbf{x},\textbf{y})$.
\ENSURE The network parameters $\bm{\theta}$ of ScasNet.
\STATE Initialize $\bm{\theta}$ and the learning rate $\eta$.
\REPEAT
\STATE Call the encoder forward pass to obtain feature maps of different levels $\mathrm{F}=$ {\sc{FeatureExtraction}}$(\textbf{x}, \bm{\theta})$.
\STATE Aggregate multi-context information $\mathrm{T}=$ {\sc{MultiScaleContextsAggregation}}$(\mathrm{F}, \bm{\theta})$ by Eq. \eqref{Eq:context}.
\STATE Perform refinement to obtain the refined feature map $f(\textbf{x}) = ${\sc{Refinement}}$(\mathrm{T}, \mathrm{F}, \bm{\theta})$ by Eq. \eqref{Eq:refine}.
\STATE Calculate $\text{Loss}({\bm \theta}) = ${\sc{NormalizedCrossEntropyLoss}}$(\textbf{y}, f(\textbf{x}))$ by Eq. \eqref{Eq:loss} and Eq. \eqref{Eq:softmax}
\STATE Calculate the back propagation gradient $\frac{\partial\, \text{Loss}({\bm \theta})}{\partial\, \bm{\theta}}$ by Eq. \eqref{Eq:derivative_f(x)} and Eq. \eqref{Eq:chain_rule} with chain rule.
\STATE Update $\bm{\theta} \leftarrow \bm{\theta}-\eta\,\frac{\partial\, \text{Loss}({\bm \theta})}{\partial\, \bm{\theta}}$.
\UNTIL {$\text{Loss}({\bm \theta})$ converges}
\RETURN $\bm{\theta}$
\end{algorithmic}
\end{algorithm}
\par}

{\renewcommand\baselinestretch{1}\selectfont
\renewcommand{\algorithmicrequire}{\textbf{Input:}}
\renewcommand{\algorithmicensure}{\textbf{Output:}}
\renewcommand{\algorithmicrepeat}{\textbf{Repeat:}}
\renewcommand{\algorithmicuntil}{\textbf{Until:}}
\renewcommand{\algorithmicreturn}{\textbf{Return:}}
\begin{algorithm}[t]
\caption{\ Inference procedure of the proposed ScasNet}
\label{alg:inference}
\small
\begin{algorithmic}[1]
\REQUIRE The image data $\textbf{x}$ and the number of scales $L$.
\ENSURE The prediction labeling map $\textbf{k}$.
\STATE Initialize the network parameters $\bm{\theta}$ outputted by Algorithm \ref{alg:learning} \\ and the average prediction probability map $p_{\textbf{k}}(\textbf{x})=\bm{0}$.
\FOR {$\ell$ in $\{1,\dots,L\}$}
\STATE Calculate the resized image $\textbf{x}^{\ell} = $ {\sc{ResizeWithBilnearInterpolation}}$(\textbf{x}, \ell)$.
\STATE Obtain the final feature map for the $\ell$-th scale $f^{\ell}(\textbf{x}^{\ell})= $ {\sc{NetworkForwardPass}}$(\textbf{x}^{\ell}, \bm{\theta})$.
\STATE Calculate the prediction probability map for the $\ell$-th scale $p^{\ell}_{\textbf{k}}(\textbf{x}^{\ell}) =$ {\sc{SoftmaxFunction}}$(f(\textbf{x}^{\ell}))$ by Eq. \eqref{Eq:softmax}.
\STATE Resize $p^{\ell}_{\textbf{k}}(\textbf{x}^{\ell})$ to original image size $p_{\textbf{k}}(\textbf{x}^{\ell}) = $ {\sc{ResizeWithBilnearInterpolation}}$(p^{\ell}_{\textbf{k}}(\textbf{x}^{\ell}), \textbf{x})$.
\STATE Add $p_{\textbf{k}}(\textbf{x}^{\ell})$ to the average prediction probability map $p_{\textbf{k}}(\textbf{x}) = p_{\textbf{k}}(\textbf{x}) + p_{\textbf{k}}(\textbf{x}^{\ell})$
\ENDFOR
\STATE Perform average operation $p_{\textbf{k}}(\textbf{x}) = \frac{1}{L}p_{\textbf{k}}(\textbf{x})$
\STATE Calculate the prediction labeling map $\hat{\textbf{k}} = \underset{\textbf{k}}{\rm{argmax}}\;p_{\textbf{k}}(\textbf{x})$ in Eq. \eqref{Eq:problem_formula}.
\RETURN $\hat{\textbf{k}}$
\end{algorithmic}
\end{algorithm}
\par}

It should be noted that due to the complicated structure, ResNet ScasNet has much difficulty to converge without BN layer. On the contrary, VGG ScasNet can converge well even though the BN layer is not used since it is relatively easy to train. In both of the two types of ScasNet, \emph{sum} fusion operation is performed for efficiency.

\subsection{\bf Learning and Inference}

In the learning stage, original VHR images and their corresponding reference images (i.e., ground truth) are used. Both of them are cropped into a number of patches, which are used as inputs to ScasNet. We use the normalized cross entropy loss as the learning objective, which is defined as
\begin{equation}
\label{Eq:loss}
\text{Loss}(y,f(x),{\bm \theta})=\frac{1}{MN}\sum\limits_{i=1}^{M}\sum\limits_{j=1}^{N}\sum\limits_{k=1}^{K}-\, \mathrm{I}(y_{i}^{j}=k) \text{log}\ p_{k}(x_{i}^{j}),
\end{equation}
where ${\bm \theta}$ represents the parameters of ScasNet; $M$ is the mini-batch size; $N$ is the number of pixels in each patch; $K$ is the number of categories; $\mathrm{I}(y=k)$ is an indicator function, it takes $1$ when $y=k$, and $0$ otherwise; $x_{i}^{j}$ is the $j$-th pixel in the $i$-th patch and $y_{i}^{j}$ is the ground truth label of $x_{i}^{j}$. Let $f(x_{i}^{j})$ denote the output of the layer before softmax (see Fig. \ref{fig:scasnet_overview}) at pixel $x_{i}^{j}$, the probability of the pixel $x_{i}^{j}$ belonging to the $k$-th category $p_{k}(x_{i}^{j})$ is defined by the softmax function, that is
\begin{equation}
\label{Eq:softmax}
p_{k}(x_{i}^{j})=\frac{\text{exp}(f_{k}(x_{i}^{j}))}{\sum\limits_{l=1}^{K}\text{exp}(f_{l}(x_{i}^{j}))}.
\end{equation}

To train ScasNet in the end-to-end manner, $\text{Loss}({\bm \theta})$ is minimized w.r.t. the ScasNet parameters ${\bm \theta}$. We have to first calculate the derivative of the loss in Eq. \eqref{Eq:loss} w.r.t. the parameters of different component layers with chain rule, and then update the parameters layer-by-layer with back propagation. For clarity, we only present the generic derivative of loss to the output of the layer before softmax and other hidden layers. The derivative of $\text{Loss}({\bm \theta})$ to the output (i.e., $f_{k}(x_{i}^{j})$) of the layer before softmax is calculated as:
\begin{equation}
\label{Eq:derivative_f(x)}
\frac{\partial\, \text{Loss}({\bm \theta})}{\partial\, f_{k}(x_{i}^{j})}=\frac{1}{MN}\sum\limits_{i=1}^{M}\sum\limits_{j=1}^{N}\sum\limits_{k=1}^{K}-\,\mathrm{I}(y_{i}^{j}=k)\left(1-p_{k}(x_{i}^{j})\right).
\end{equation}
The specific derivation process can be referred in the Appendix A of supplementary material.

The derivative of $\text{Loss}({\bm \theta})$ to each hidden (i.e., $h_{k}(x_{i}^{j})$) layer can be obtained with the chain rule as:
\begin{equation}
\begin{aligned}
\label{Eq:chain_rule}
\frac{\partial\, \text{Loss}({\bm \theta})}{\partial\, h_{k}(x_{i}^{j})}=\frac{\partial\, \text{Loss}({\bm \theta})}{\partial\, f_{k}(x_{i}^{j})}\frac{\partial\, f_{k}(x_{i}^{j})}{\partial\, h_{k}(x_{i}^{j})}.
\end{aligned}
\end{equation}
The first item in Eq. \eqref{Eq:chain_rule} is given in Eq. \eqref{Eq:derivative_f(x)}, and the second item also can be obtained by corresponding chain rule.

\renewcommand{\thefootnote}{\arabic{footnote}}
\setcounter{footnote}{0}
The pseudo-code of learning procedure of ScasNet is shown in Algorithm \ref{alg:learning}. In the experiments, we implement ScasNet based on the Caffe framework \citep{conf_mm_JiaSDKLGGD14}. The image patches of size $400\times400$ are used as inputs \footnote{The possibly few number of categories in these patches doesn't influence the high diversity of categories in raw VHR images.}. Due to the limit of GPU memory, we set the mini-batch size as $4$. To train ScasNet, we use stochastic gradient descent (SGD) with initial learning rate of $0.01$, and drop the learning rate by a factor of $0.1$ every $20$ epochs. The momentum and weight decay are set as $0.9$ and $0.0005$, respectively. Experimentally, ScasNet is trained for about $80$ epochs.
\begin{table*}[t]
  \centering
  \scriptsize
  \caption{The detailed information of experimental setting on the three datasets. `offline/online' denotes the training set for offline validation and the training set for online test, respectively.}
  \label{tab:data_description}
  \begin{tabular}{r||c|c|c|c}
  \Xhline{1.5pt}
  \multirow{2}*{Dataset}&\multicolumn{2}{c|}{TRAINING SET}&VALIDATION SET&TEST SET\\
  \cline{2-5}
  &Images&Patches (400$\times$400)&Images&Images\\
  \Xhline{0.8pt}
  Massachusetts Building& 141 & 20727 & $0$ & $10$ \\
  \hline
  \multirow{2}{*}{Vaihingen Challenge} & offline$/$online & offline$/$online & \multirow{2}{*}{$8$} & \multirow{2}{*}{$17$} \\
  \cline{2-3}
   & $8/16$ & $12384/24400$ & & \\
  \hline
  \multirow{2}{*}{Potsdam Challenge} & offline$/$online & offline$/$online & \multirow{2}{*}{$10$} & \multirow{2}{*}{$14$} \\
  \cline{2-3}
   & $14/24$ & $16800/28800$ & & \\
  \Xhline{1.5pt}
  \end{tabular}
\end{table*}
\begin{figure*}[t]
\centerline{\includegraphics[width=18.1cm]{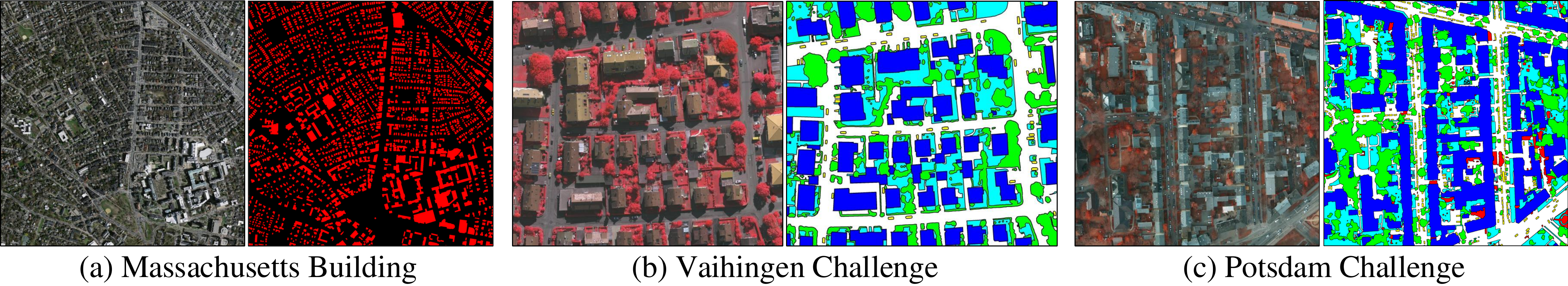}}
\caption{The image samples and corresponding ground truth on the three datasets. The label of Massachusetts building includes two categories: building (red) and background (black). The label of Vaihingen and Potsdam challenge includes six categories: impervious surface (imp surf, white), building (blue), low vegetation (low veg, cyan), tree (green), car (yellow) and clutter/background (red), where the boundary (black) is depicted for visual clarity.}
\label{fig:datashow}
\end{figure*}

The pseudo-code of inference procedure is shown in Algorithm \ref{alg:inference}. In the inference stage, we perform multi-scale inference of 0.5, 1 and 1.5 times the size of raw images (i.e., $L=3$ scales), and we average the final outputs at all the three scales. Specifically, we first crop a resized image (i.e., $\textbf{x}^{\ell}$) into a series of patches without overlap. Then, the prediction probability maps of these patches are predicted by inputting them into ScasNet with a forward pass. Finally, the entire prediction probability map (i.e., $p^{\ell}_{\textbf{k}}(\textbf{x}^{\ell})$) of this image is constituted by the probability maps of all patches. The purpose of multi-scale inference is to mitigate the discontinuity in final labeling map caused by the interrupts between patches.

\section{Experiments and Evaluations}
\label{Section4}
In this section, dataset description, experimental setting, comparing methods and extensive experiments in both qualitative and quantitative comparisons are first presented. Then, the proposed ScasNet is analyzed in detail by a series of ablation experiments.

\subsection{\bf Dataset Description}

We evaluate the proposed ScasNet on three challenging public datasets for semantic labeling.
\begin{figure*}[t]
\centerline{\includegraphics[width=18.1cm]{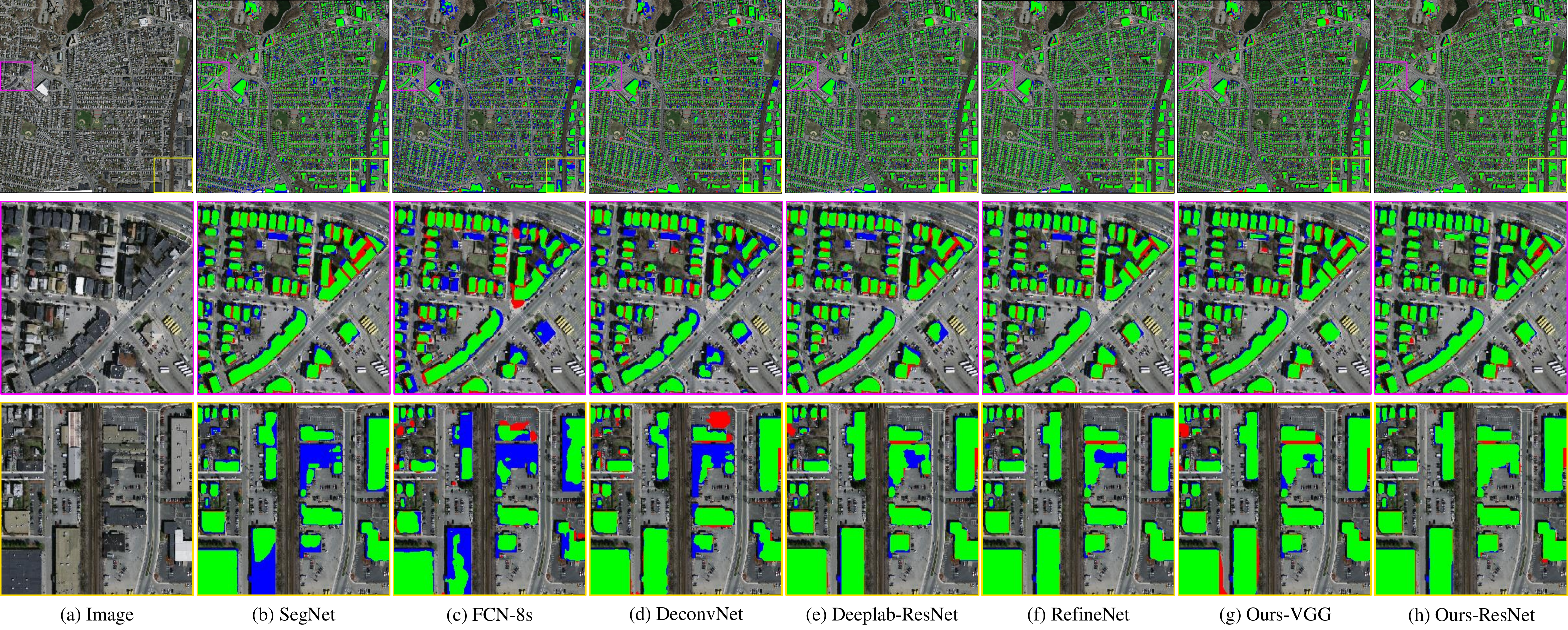}}
\caption{Qualitative comparison with the state-of-the-art deep models on Massachusetts building TEST SET. The 1st row illustrates the overall results of one image sample, and the last two rows show the close-ups of the corresponding regions in the 1st row. In the colored figures, true positive (tp) is marked in green, false positive (fp) in red and false negative (fn) in blue. (Best viewed in color)}
\label{fig:compare_mnih}
\end{figure*}

{\bf Massachusetts Building Dataset:} This dataset is proposed by Mnih \citep{thesis_mnih2013}. It consists of $151$ aerial images of the Boston area, with each of the images being $1500\times1500$ pixels at a GSD (Ground Sampling Distance) of $1$m. The ground truth of all these images are available. We randomly split the data into a training set of 141 images, and a test set of 10 images. As Fig. \ref{fig:datashow}(a) shows, it covers mostly urban areas and buildings of all sizes, including houses and garages.

{\bf ISPRS Vaihingen Challenge Dataset:} This is a benchmark dataset for \emph{ISPRS 2D Semantic labeling challenge} in Vaihingen \citep{ISPRS_challenge}. It consists of $3$-band IRRG (Infrared, Red and Green) image data, and corresponding DSM (Digital Surface Model) and NDSM (Normalized Digital Surface Model) data. Overall, there are $33$ images of $\approx2500\times2000$ pixels at a GSD of $\approx9$cm in image data. Among them, the ground truth of only $16$ images are available, and those of the remaining $17$ images are withheld by the challenge organizer for online test. For offline validation, we randomly split the $16$ images with ground truth available into a training set of $8$ images, and a validation set of $8$ images. For online test, we use all the $16$ images as training set. Note that DSM and NDSM data in all the experiments on this dataset are not used.

{\bf ISPRS Potsdam Challenge Dataset:} This is a benchmark dataset for \emph{ISPRS 2D Semantic labeling challenge} in Potsdam \citep{ISPRS_challenge}. It consists of $4$-band IRRGB (Infrared, Red, Green, Blue) image data, and corresponding DSM and NDSM data. Overall, there are $38$ images of $6000\times6000$ pixels at a GSD of $\approx5$cm. Among them, the ground truth of only $24$ images are available, and those of the remaining $14$ images are withheld by the challenge organizer for online test. For offline validation, we randomly split the $24$ images with ground truth available into a training set of $14$ images, a validation set of $10$ images. For online test, we use all the $24$ images as training set. Note that only the $3$-band IRRG images extracted from raw $4$-band data are used, and DSM and NDSM data in all the experiments on this dataset are not used.

Table \ref{tab:data_description} summarizes the detailed information of all the above datasets.  Fig. \ref{fig:datashow} shows some image samples and the ground truth on the three datasets. As it shows, there are many confusing manmade objects and intricate fine-structured objects in these VHR images, which poses much challenge for achieving both coherent and accurate semantic labeling.

\subsection{\bf Experimental Setting}

The remote sensing datasets are relatively small to train the proposed deep ScasNet. To reduce overfitting and train an effective model, data augmentation, \emph{transfer learning} \citep{conf_nips_JasonJYH2014,conf_cvpr_PenaANKSJ2015,journals_rs_FanHu15,journals_corr_XieNMLDES2015} and regularization techniques are applied.
%
%
\begin{figure}[t]
\begin{minipage}[htbp]{0.5\linewidth}
  \centering
  \scriptsize
  \caption{Quantitative comparison (\%) with the state-of-the-art deep models on Massachusetts building TEST SET, where the values in bold are the best and the values underlined are the second best.}
  \label{tab:Mnih_comparison_models}
  \begin{tabular}{r||cc}
  \Xhline{1.5pt}
  Method&IoU&F1 \\
  \hline
  SegNet&56.38&72.11 \\
  FCN-8s&50.94&67.96 \\
  DeconvNet&60.74&75.57 \\
  Deeplab-ResNet&69.50&82.01 \\
  RefineNet&\underline{71.92}&\underline{83.67} \\
  \Xhline{0.8pt}
  Ours-VGG&69.22&81.81 \\
  Ours-ResNet&\textbf{74.34}&\textbf{85.58} \\
  \Xhline{1.5pt}
  \end{tabular}
\end{minipage}
\begin{minipage}[htbp]{0.5\linewidth}
\centerline{\includegraphics[width=6cm]{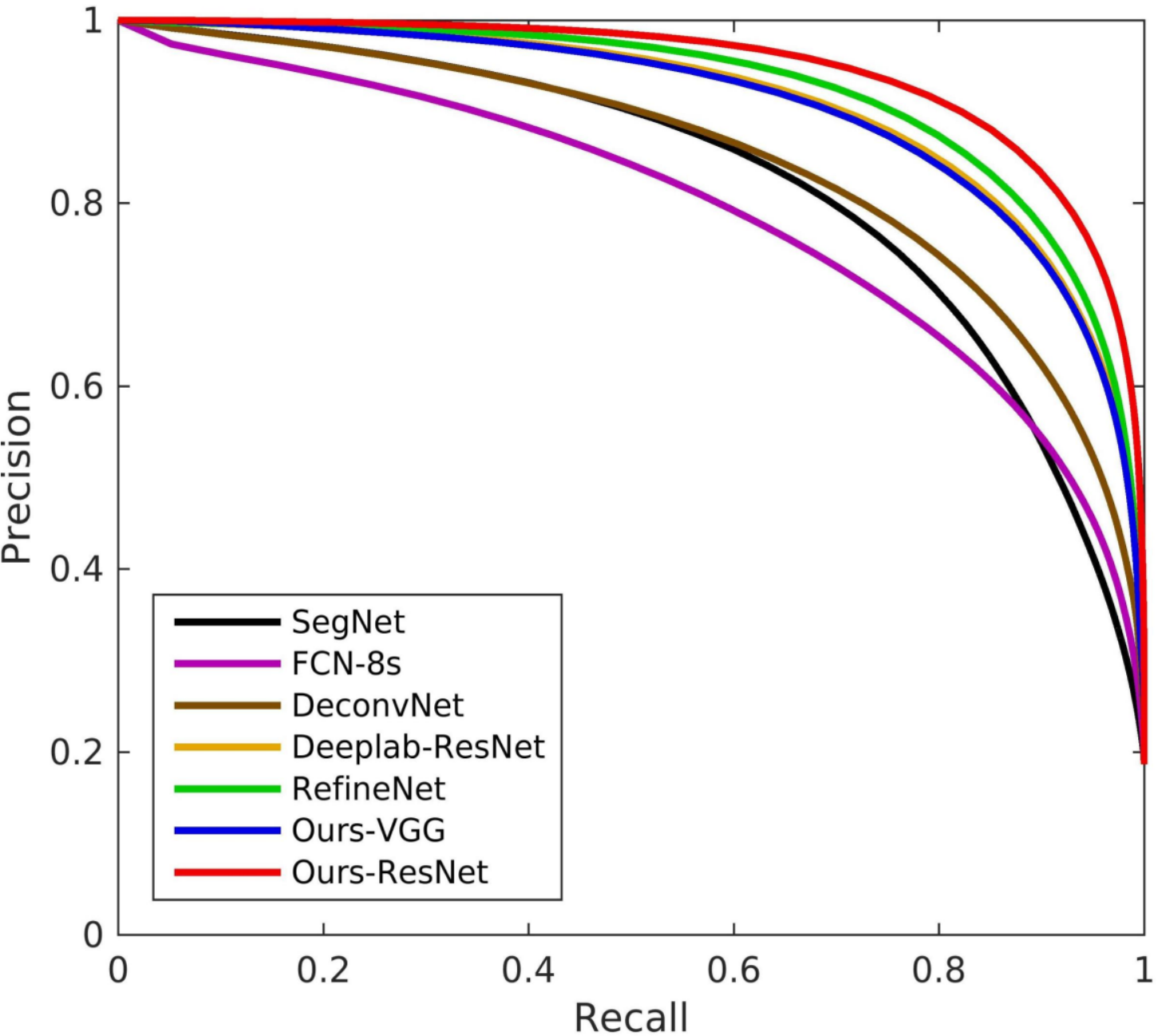}}
\caption{Precision-recall (PR) curves of all the comparing deep models on Massachusetts building TEST SET. (Best viewed in color)}
\label{fig:pre_mnih}
\end{minipage}
\end{figure}
%

In the experiments, $400\times400$ patches cropped from raw images are employed to train ScasNet. For the training sets, we use a two-stage method to perform data augmentation. In the first stage, given an image, we crop it to generate a series of $400\times400$ patches with the overlap of $100$ pixels. In the second stage, for each patch, we flip it in horizontal and vertical reflections and rotate it counterclockwise at the step of $90^{\circ}$. The detailed number of patches in the augmented data is presented in Tabel \ref{tab:data_description}.

In the experiments, the parameters of the encoder part (see Fig. \ref{fig:scasnet_overview}) in our models are initialized with the models pre-trained on PASCAL VOC 2012 \citep{journals_ijcv_EveringhamEGWWZ15}. All the other parameters in our models are initialized using the techniques introduced by He et al. \citep{conf_iccv_HeZRS15}.

To avoid overfitting, dropout technique \citep{journals_jmlr_Sriva2014} with ratio of $50\%$ is used in ScasNet, which provides a computationally inexpensive yet powerful regularization to the network.

\subsection{\bf Comparing methods}
To verify the performance, the proposed ScasNet is compared with extensive state-of-the-art methods on two aspects: deep models comparison and benchmark test comparison.
\begin{figure*}[t]
\centerline{\includegraphics[width=18.1cm]{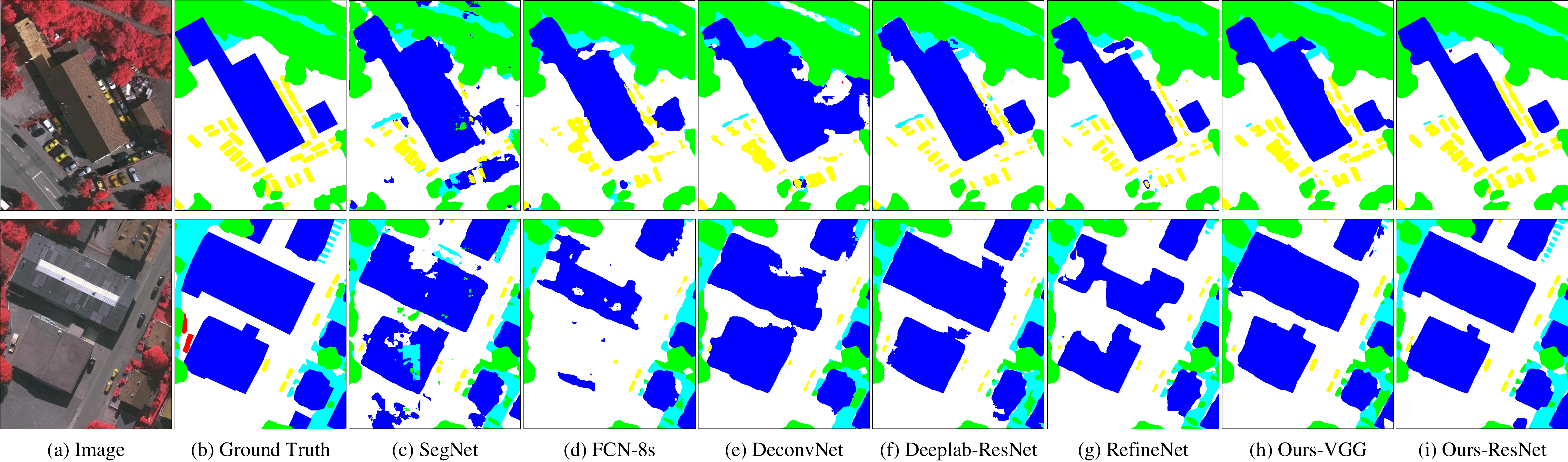}}
\caption{Qualitative comparison with the state-of-the-art deep models on \emph{ISPRS Vaihingen challenge} OFFLINE VALIDATION SET. The label includes six categories: impervious surface (imp surf, white), building (blue), low vegetation (low veg, cyan), tree (green), car (yellow) and clutter/background (red).}
\label{fig:compare_vai_offline}
\end{figure*}

{\bf Comparing Deep Models:} ScasNet is compared with five state-of-the-art deep models on the three datasets. The main information of these models (including our models) is summarized as follows:

\begin{enumerate}[1)]
\item Ours-VGG: The self-cascaded network with the encoder based on a variant of $16$-layer VGG-Net \citep{conf_iclr_ChenPKMY15}.
\item Ours-ResNet: The self-cascaded network with the encoder based on a variant of $101$-layer ResNet \citep{journals_corr_ZhaoSQWJ2016}.
\item {FCN-8s}: Long et al. \citep{conf_cvpr_LongSD15} propose FCN for semantic segmentation, which achieves the state-of-the-art performance on three benchmarks \citep{journals_ijcv_EveringhamEGWWZ15,conf_NYUDv2_NatDPR2012,conf_eccv_Liu2008}. There are three versions of FCN models: FCN-$32$s, FCN-$16$s and FCN-$8$s. We use the best performance model FCN-$8$s as comparison.
\item {SegNet}: Badrinarayanan et al. \citep{journals_corr_BadrinarayananK15} propose SegNet for semantic segmentation of road scene, in which the decoder uses pooling indices in the encoder to perform non-linear \emph{up-sampling}. It provides competitive performance while works faster than most of the other models.
\item {DconvNet}: Deconvolutional network (DconvNet) is proposed by Noh et al. \citep{conf_iccv_NohHH15} for semantic segmentation, which is composed of \emph{deconvolution} and \emph{un-pooling} layers. It achieves the state-of-the-art performance on PASCAL VOC 2012 \citep{journals_ijcv_EveringhamEGWWZ15}.
\item {Deeplab-ResNet}: Chen et al. \citep{conf_iclr_ChenPKMY15} propose Deeplab-ResNet based on three $101$-layer ResNet \citep{conf_cvpr_HeZRS2016}, which achieves the state-of-the-art performance on PASCAL VOC 2012 \citep{journals_ijcv_EveringhamEGWWZ15}. Actually, they use three-scale (0.5, 0.75 and 1 the size of input image) images as input to three 101-layer ResNet respectively, and then fuse three outputs as final prediction.
\item {RefineNet}: RefineNet is proposed by Lin et al. \citep{journals_corr_LinMSR2016} for semantic segmentation, which is based on ResNet \citep{conf_cvpr_HeZRS2016}. It achieves the state-of-the-art performance on seven benchmarks, such as PASCAL VOC 2012 \citep{journals_ijcv_EveringhamEGWWZ15} and NYUDv2\citep{conf_NYUDv2_NatDPR2012}. Here, we take RefineNet based on $101$-layer ResNet for comparison.
\end{enumerate}

It should be noted that all the experimental settings for the above models are the same, except for two aspects. Firstly, their training hyper-parameter values used in the Caffe framework \citep{conf_mm_JiaSDKLGGD14} are different. This is because it may need different hyper-parameter values (such as learning rate) to make them converge when training different deep models. Secondly, all the models are trained based on the widely used \emph{transfer learning} \citep{conf_nips_JasonJYH2014,conf_cvpr_PenaANKSJ2015,journals_rs_FanHu15,journals_corr_XieNMLDES2015} in the field of \emph{deep learning}. Specifically, except for our models, all the other models are trained by finetuning their corresponding best models pre-trained on PASCAL VOC 2012 \citep{journals_ijcv_EveringhamEGWWZ15} on semantic segmentation task. For our models, only the parameters of the encoder part (see Fig. \ref{fig:scasnet_overview}) are initialized with the pre-trained models. Furthermore, the influence of \emph{transfer learning} on our models is analyzed in Section \ref{subsection:model_analysis}.
\begin{table*}[t]
  \centering
  \scriptsize
  \caption{Quantitative comparison (\%) with the state-of-the-art deep models on \emph{ISPRS Vaihingen challenge} OFFLINE VALIDATION SET, where the values in bold are the best and the values underlined are the second best.}
  \label{tab:Vai_comparison_models}
  \begin{tabular}{r||cc|cc|cc|cc|cc|cc}
  \Xhline{1.5pt}

  \multirow{2}*{Model}&\multicolumn{2}{c|}{imp surf}&\multicolumn{2}{c|}{building}&\multicolumn{2}{c|}{low veg}&
  \multicolumn{2}{c|}{tree}&\multicolumn{2}{c|}{car}&\multicolumn{2}{c}{Avg.}\\
  \cline{2-13}
  &IoU&F1&IoU&F1&IoU&F1&IoU&F1&IoU&F1&mean IoU&mean F1 \\
  \Xhline{0.8pt}
  SegNet&66.85&80.13&76.10&86.43&50.56&68.65&69.71&82.15&62.38&76.83&65.12&78.83 \\
  FCN-8s&75.26&85.89&80.51&89.20&65.58&79.21&70.49&82.69&45.84&62.87&67.54&79.97 \\
  DeconvNet&80.27&89.06&87.19&93.16&68.57&81.36&74.91&85.65&51.93&68.36&72.57&83.52 \\
  Deeplab-ResNet&82.20&90.23&\underline{91.22}&\underline{95.41}&\underline{71.12}&\underline{83.12}&
  \underline{76.93}&\underline{86.96}&56.78&72.43&75.65&85.63 \\
  RefineNet&80.08&88.94&88.62&93.97&70.69&82.83&76.00&86.36&68.35&81.56&76.75&86.73 \\
  \Xhline{0.8pt}
  Ours-VGG&\underline{82.70}&\underline{90.53}&89.54&94.48&69.00&81.66&76.17&86.47&\underline{76.89}&\underline{86.93}&\underline{78.86}&\underline{88.02} \\
  Ours-ResNet&\textbf{85.86}&\textbf{93.76}&\textbf{92.45}&\textbf{96.19}&\textbf{76.26}&\textbf{87.62}&
  \textbf{83.77}&\textbf{90.61}&\textbf{81.14}&\textbf{89.81}&\textbf{83.90}&\textbf{91.60} \\

  \Xhline{1.5pt}
  \end{tabular}
\end{table*}
\begin{figure*}[t]
\centerline{\includegraphics[width=18.1cm]{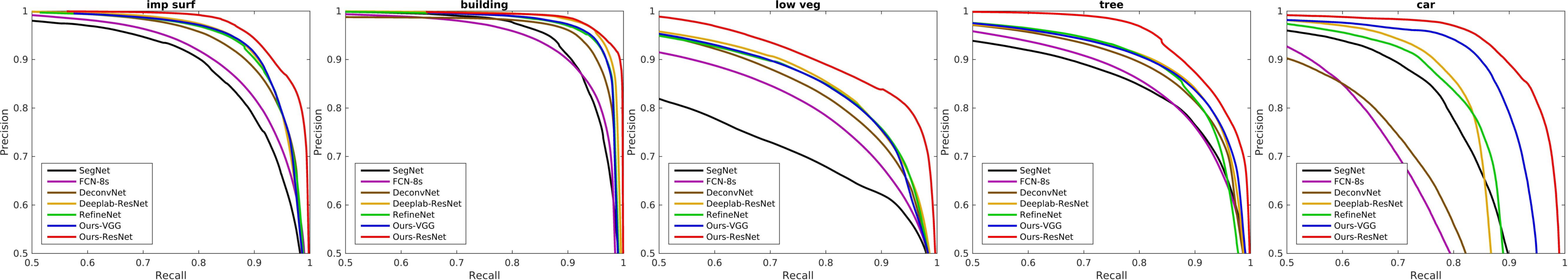}}
\caption{Precision-recall (PR) curves of all the comparing deep models on \emph{ISPRS Vaihingen challenge} OFFLINE VALIDATION SET. Categories from left to right: impervious surface (imp surf), building, low vegetation (low veg), tree , car. (Best viewed in color)}
\label{fig:pre_rec_vai}
\end{figure*}

\renewcommand{\thefootnote}{\fnsymbol{footnote}}
\setcounter{footnote}{0}
{\bf Benchmark Comparing Methods:} By submitting the results of test set to the \emph{ISPRS challenge} organizer, ScasNet is also compared with other competitors' methods on benchmark test. The details of these methods (including our methods) are listed as follows, where the names in brackets are the short names on the challenge evaluation website \footnote{\url{http://www2.isprs.org/commissions/comm3/wg4/results.html}}:
\begin{enumerate}[1)]
\item {Ours-ResNet (\textbf{`CASIA2'})}: The single self-cascaded network with the encoder based on a variant of $101$-layer ResNet \citep{journals_corr_ZhaoSQWJ2016}. In our method, only raw image data is used for training. Specifically, 3-band IRRG images are used for Vaihingen and only 3-band IRRG images obtained from raw image data (i.e., 4-band IRRGB images) are used for Potsdam. Moreover, we do not use the elevation data (DSM and NDSM), additional hand-crafted features, model ensemble strategy or any postprocessing.
\item {SVL-features + DSM + Boosting + CRF (\textbf{`SVL\_*'})}: The method as baseline implemented by the challenge organizer \citep{report_Markus2015}. In addition to the standard SVL-features \citep{book_SVL2011}, they also use NDVI (Normalized Digital Vegetation Index), saturation and NDSM features. Then, an Adaboost-based classifier is trained. A CRF (Conditional Random Field) model is applied to obtain final prediction. For comparison, `SVL\_6' is compared for Vaihingen and `SVL\_3' (no CRF) for Potsdam.
\item {CNN + NDSM + Deconvolution (\textbf{`UZ\_1'})}: The method proposed by \citep{journals_tgrs_VolMTD2017}. They use an downsample-then-upsample architecture , in which rough spatial maps are first learned by convolutions and then these maps are upsampled by \emph{deconvolution}. NDSM data is used in their method.
\item {CNN + DSM + NDSM + RF + CRF (\textbf{`ADL\_3'})}: The method proposed by \citep{journals_ijeors_paiSSHA2016}. They apply both CNN and hand-crafted features to dense image patches to produce per-pixel category probabilities. Random forest (RF) classifier is trained on hand-crafted features and the output probabilities are combined with those generated by the CNN. CRF is applied as a postprocessing step.
\item {FCN + DSM + RF + CRF (\textbf{`DST\_2'})}: The method proposed by \citep{journals_corr_sheJ2016}. They use a hybrid FCN architecture to combine image data with DSM data. Then, CRF is applied as a postprocessing step.
\item {FCN + SegNet + VGG + DSM + Edge (\textbf{`DLR\_8'})}: The method proposed by \citep{journals_corr_MarSWJSU2016}. They use a multi-scale ensemble of FCN, SegNet and VGG, incorporating both image data and DSM data. Moreover, they combine semantic labeling with informed edge detection.
\item {SegNet + DSM + NDSM (\textbf{`ONE\_7'})}: The method proposed by \citep{journals_corr_AudebSLLS2016}. They fuse the output of two multi-scale SegNets, which are trained with IRRG images and synthetic data (NDVI, DSM and NDSM) respectively.
\item {CNN + DSM + SVM (\textbf{`GU'})}: In their method, both image data and DSM data are used to train a CNN. Moreover, CNN is trained on six scales of the input data. Finally, a SVM maps the six predictions into a single-label.
\item {CNN + DSM (\textbf{`AZ\_1'})}: In their method, a CNN with encoder-decoder architecture is used. The input to the network includes six channels of IRRGB, NDVI, and NDSM, which are concatenated together.
\item {SegNet + NDSM (\textbf{`RIT\_2'})}: In their method, two SegNets are trained with RGB images and synthetic data (IR, NDVI and NDSM) respectively. Then, feature fusion in the early stages is performed.
\end{enumerate}

\begin{figure*}[t]
\centerline{\includegraphics[width=18.1cm]{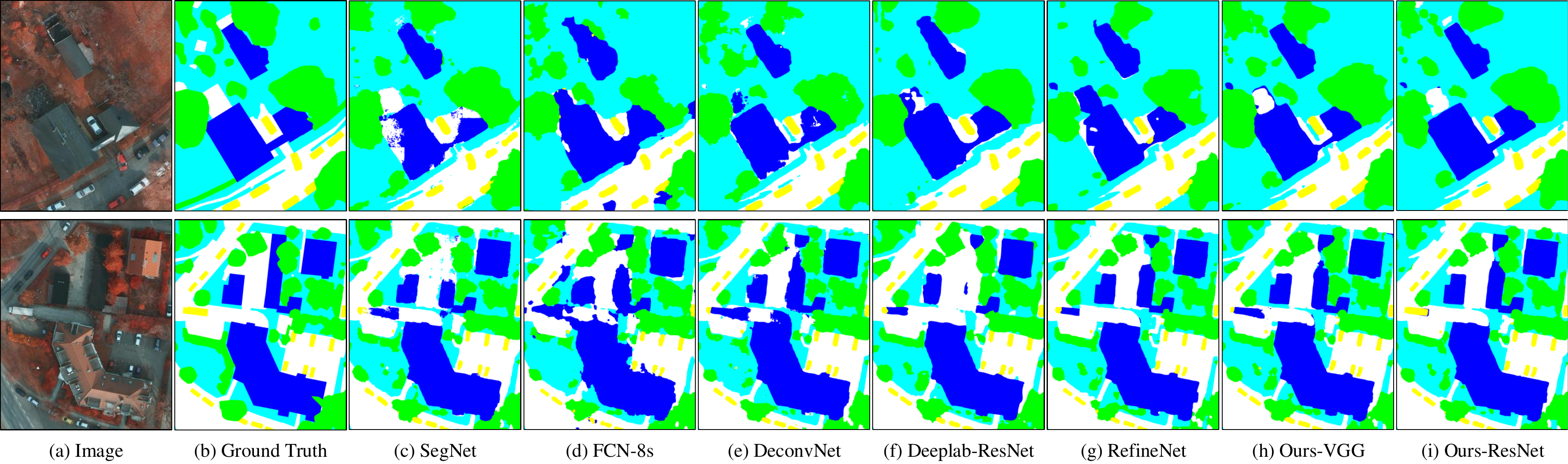}}
\caption{Qualitative comparison with the state-of-the-art deep models on \emph{ISPRS Potsdam challenge} OFFLINE VALIDATION SET. The label includes six categories: impervious surface (imp surf, white), building (blue), low vegetation (low veg, cyan), tree (green), car (yellow) and clutter/background (red).}
\label{fig:compare_pot_offline}
\end{figure*}
\subsection{\bf  Evaluation Metrics}

To assess the quantitative performance, two overall benchmark metrics are used, i.e., \emph{F1 score} (F1) and \emph{intersection over union} (IoU). F1 is defined as
\begin{equation}
\label{Eq:F1}
  \textrm{F1} = 2\, \frac{\mathrm{Pre} \times \mathrm{Rec}}{\mathrm{Pre} + \mathrm{Rec}}, \, \textrm{Pre} = \frac{tp}{tp+fp}, \, \textrm{Rec} = \frac{tp}{tp+fn}.
\end{equation}
Here, $tp$, $fp$ and $fn$ are the number of true positives, false positives and false negatives, respectively.

IoU is defined as:
\begin{equation}
\label{Eq:IoU}
  \textrm{IoU}(\mathcal{P}_m, \mathcal{P}_{gt}) = \frac{|\mathcal{P}_m \cap \mathcal{P}_{gt}|}{|\mathcal{P}_m \cup \mathcal{P}_{gt}|},
\end{equation}
where $\mathcal{P}_{gt}$ is the set of ground truth pixels and $\mathcal{P}_{m}$ is the set of prediction pixels, `$\cap$' and `$\cup$' denote \emph{intersection} and \emph{union} operations, respectively. $|\bm \cdot|$ denotes calculating the number of pixels in the set.
\begin{table*}[t]
  \centering
  \scriptsize
  \caption{Quantitative comparison (\%) with the state-of-the-art deep models on \emph{ISPRS Potsdam challenge} OFFLINE VALIDATION SET, where the values in bold are the best and the values underlined are the second best.}
  \label{tab:Pot_comparison_models}
  \begin{tabular}{r||cc|cc|cc|cc|cc|cc}
  \Xhline{1.5pt}

  \multirow{2}*{Model}&\multicolumn{2}{c|}{imp surf}&\multicolumn{2}{c|}{building}&\multicolumn{2}{c|}{low veg}&
  \multicolumn{2}{c|}{tree}&\multicolumn{2}{c|}{car}&\multicolumn{2}{c}{Avg.}\\
  \cline{2-13}
  &IoU&F1&IoU&F1&IoU&F1&IoU&F1&IoU&F1&mean IoU&mean F1 \\
  \Xhline{0.8pt}
  SegNet&85.42&92.14&91.17&95.38&78.58&88.01&75.71&86.17&88.12&93.68&83.80&91.08 \\
  FCN-8s&77.55&87.35&79.94&88.85&71.95&83.68&69.53&82.02&79.68&88.69&75.73&86.12 \\
  DeconvNet&87.08&93.09&93.12&96.44&77.59&87.38&71.67&83.50&92.28&95.98&84.35&91.28 \\
  Deeplab-ResNet&88.23&93.75&\underline{94.39}&\underline{97.11}&78.85&88.18&74.50&85.39&87.11&93.11&84.62&91.51 \\
  RefineNet&86.80&92.93&91.13&95.36&78.69&88.07&73.51&84.74&92.75&96.24&84.58&91.47 \\
  \Xhline{0.8pt}
  Ours-VGG&\underline{88.68}&\underline{94.00}&94.12&96.97&\underline{80.67}&\underline{89.30}&\textbf{77.86}&\textbf{87.55}&\underline{94.07}&\underline{96.94}&\underline{87.08}&\underline{92.95} \\
  Ours-ResNet&\textbf{90.06}&\textbf{94.77}&\textbf{96.27}&\textbf{98.10}&\textbf{80.83}&\textbf{89.40}&\underline{76.86}&\underline{86.92}&\textbf{94.90}&\textbf{97.38}&\textbf{87.78}&\textbf{93.31} \\

  \Xhline{1.5pt}
  \end{tabular}
\end{table*}
\begin{figure*}[t]
\centerline{\includegraphics[width=18.1cm]{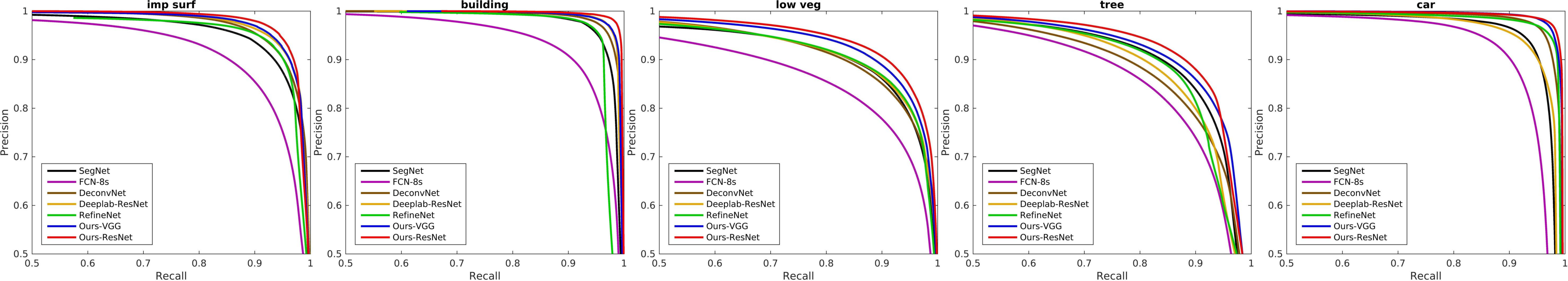}}
\caption{Precision-recall (PR) curves of all the comparing deep models on \emph{ISPRS Potsdam challenge} OFFLINE VALIDATION SET. Categories from left to right: impervious surface (imp surf), building, low vegetation (low veg), tree , car. (Best viewed in color)}
\label{fig:pre_pot}
\end{figure*}

To evaluate the performance of different comparing deep models, we compare the above two metrics on each category, and the mean value of metrics to assess the average performance. Furthermore, precision-recall (PR) curve is drawn to qualify the relation between \emph{precision} and \emph{recall} on each category. Specifically, the predicted score maps are first binarized using different thresholds varying from $0$ to $1$. Then by comparing these binarized results with the ground truth, a series of precision-recall values can be obtained to plot the PR curve.

When compared with other competitors' methods on benchmark test \citep{ISPRS_challenge}, besides the F1 metric for each category, the \emph{overall accuracy} (Overall Acc.) derived from the pixel-based confusion matrix \citep{ISPRS_challenge} is also compared to assess the global performance.

It should be noted that all the metrics are computed using an alternative ground truth in which the boundaries of objects have been eroded by a $3$-pixel radius. The eroded areas are ignored during evaluation, so as to reduce the impact of uncertain border definitions.

\subsection{\bf Comparison with Deep Models}
To evaluate the effectiveness of the proposed ScasNet, the comparisons with five state-of-the-art deep models on the three challenging datasets are presented as follows:
\begin{enumerate}[1)]
\item {\bf{Massachusetts Building Test Set}}: As the global visual performance (see the 1st row in Fig. \ref{fig:compare_mnih}) and local close-ups (see the last two rows in Fig. \ref{fig:compare_mnih}) show, SegNet, FCN-8s and DeconvNet have difficulty in recognizing confusing size-varied buildings. For fine-structured buildings, FCN-8s performs incomplete and inaccurate labeling while SegNet and DeconvNet do better. The results of Deeplab-ResNet, RefineNet and Ours-VGG are relatively good, but they tend to have more false negatives (blue). Ours-ResNet generates more coherent labeling on both confusing and fine-structured buildings. Table \ref{tab:Mnih_comparison_models} summarizes the quantitative performance. As it shows, Ours-VGG achieves almost the same performance with Deeplab-ResNet, while Ours-ResNet achieves more decent score. Fig. \ref{fig:pre_mnih} shows the PR curves of all the deep models, in which both Our-VGG and Our-ResNet achieve superior performances.
    \begin{table*}[t]
  \centering
  \scriptsize
  \caption{Quantitative comparison (\%) with other competitors' methods on \emph{ISPRS Vaihingen challenge} ONLINE TEST SET, where the values in bold are the best and the values underlined are the second best. The names in brackets are the short names on the
  challenge evaluation website.}
  \label{tab:Vai_comparison_online}
  \begin{tabular}{r||c|c|c|c|c|c}
  \Xhline{1.5pt}

  Method&imp surf&building&low veg&tree&car&Overall Acc. \\
  \cline{2-7}
  \Xhline{0.7pt}
  SVL-features + DSM + Boosting + CRF ({\bf `SVL\_6'})&86.00&90.20&75.60&82.10&45.40&83.20 \\
  CNN + NDSM + Deconvolution ({\bf `UZ\_1'})&89.20&92.50&81.60&86.90&57.30&87.30 \\
  CNN + DSM + NDSM + RF + CRF ({\bf `ADL\_3'})&89.50&93.20&82.30&88.20&63.30&88.00 \\
  FCN + DSM + RF + CRF ({\bf `DST\_2'})&90.50&93.70&83.40&89.20&72.60&89.10 \\
  FCN + SegNet + VGG + DSM + Edge ({\bf `DLR\_8'})&90.40&93.60&83.90&\underline{89.70}&76.90&89.20 \\
  SegNet + DSM + NDSM ({\bf `ONE\_7'})&\underline{91.00}&\underline{94.50}&\underline{84.40}&\textbf{89.90}&\underline{77.80}&\underline{89.80} \\
  \Xhline{0.7pt}
  Ours-ResNet ({\bf `CASIA2'})&\textbf{93.20}&\textbf{96.00}&\textbf{84.70}&\textbf{89.90}&\textbf{86.70}&\textbf{91.10} \\

  \Xhline{1.5pt}
  \end{tabular}
\end{table*}

\begin{table*}[t]
  \centering
  \scriptsize
 \caption{Quantitative comparison (\%) with other competitors' methods on \emph{ISPRS Potsdam challenge} ONLINE TEST SET, where the values in bold are the best and the values underlined are the second best. The names in brackets are the short names on the
  challenge evaluation website.}
  \label{tab:Pot_comparison_online}
  \begin{tabular}{r||c|c|c|c|c|c}
  \Xhline{1.5pt}

  Method&imp surf&building&low veg&tree&car&Overall Acc. \\
  \cline{2-7}
  \Xhline{0.7pt}
  SVL-features + DSM + Boosting ({\bf `SVL\_3'})&84.00&89.80&72.00&59.00&69.80&77.20 \\
  CNN + DSM + SVM ({\bf `GU'})&87.10&94.70&77.10&73.90&81.20&82.90 \\
  CNN + NDSM + Deconvolution ({\bf `UZ\_1'})&89.30&95.40&81.80&80.50&86.50&85.80 \\
  CNN + DSM ({\bf `AZ\_1'})&91.40&96.10&86.10&86.60&93.30&89.20 \\
  SegNet + NDSM ({\bf `RIT\_2'})&\underline{92.00}&\underline{96.30}&85.50&86.50&\underline{94.50}&89.40 \\
  FCN + DSM + RF + CRF ({\bf `DST\_2'})&91.80&95.90&\underline{86.30}&\underline{87.70}&89.20&\underline{89.70} \\
  \Xhline{0.7pt}
  Ours-ResNet ({\bf `CASIA2'})&\textbf{93.30}&\textbf{97.00}&\textbf{87.70}&\textbf{88.40}&\textbf{96.20}&\textbf{91.10} \\

  \Xhline{1.5pt}
  \end{tabular}
\end{table*}
\item {\bf{Vaihingen Challenge Validation Set}}: As shown in Fig. \ref{fig:compare_vai_offline}, SegNet, FCN-8s, DeconvNet and RefineNet are sensitive to the cast shadows of buildings and trees. They can not distinguish similar manmade objects well, such as buildings and roads. Meanwhile, for fine-structured objects, these methods tend to obtain inaccurate localization, especially for the car. The results of Deeplab-ResNet are relatively coherent, while they are still less accurate. Ours-VGG and Ours-ResNet show better robustness to the cast shadows. They can achieve coherent labeling for confusing manmade objects. Moreover, fine-structured objects also can be labeled with precise localization using our models. The quantitative performance is shown in Table \ref{tab:Vai_comparison_models}. As can be seen, the performance of our best model outperforms other advanced models by a considerable margin on each category, especially for the car. Furthermore, the PR curves shown in Fig. \ref{fig:pre_rec_vai} exhibit that, our best model performs better on all the given categories.
\item {\bf{Potsdam Challenge Validation Set}}: As Fig. \ref{fig:compare_pot_offline} shows, all the five comparing models are less effective in the recognition of confusing manmade objects. They are not robust enough to the occlusions and cast shadows. For fine-structured objects like the car, FCN-8s performs less accurate localization, while other four models do better. Although the labeling results of our models have a few flaws, they can achieve relatively more coherent labeling and more precise boundaries. Table \ref{tab:Pot_comparison_models} summarizes the quantitative performance. As it shows, in labeling the VHR images with such a high resolution of $5$cm, all these models achieve decent results. Still, the performance of our best model exceeds other advanced models by a considerable margin, especially for the car. Moreover, as the PR curves in Fig. \ref{fig:pre_pot} show, our best model presents very decent performance.
\end{enumerate}

In short, the above comparisons show that, on one hand, the proposed ScasNet has strong recognition ability for confusing manmade objects in VHR images. Meanwhile, ScasNet is quite robust to the occlusions and cast shadows, and it can perform coherent labeling even for very uneven regions. These results demonstrate the effectiveness of our multi-scale contexts aggregation approach. On the other hand, ScasNet can label size-varied objects completely, resulting in accurate and smooth results, especially for the fine-structured objects like the car. This demonstrates the validity of our refinement strategy.

\subsection{\bf Comparison on Benchmark Test}
To further evaluate the effectiveness of the proposed ScasNet, comparisons with other competitors' methods on the two challenging benchmarks are presented as follows:

\begin{table*}[t]
  \centering
  \scriptsize
  \caption{Quantitative comparison (\%) between 3-scale test (0.5, 1 and 1.5 times the size of raw image) and 1-scale test on \emph{ISPRS Vaihingen \& Potsdam challenge} ONLINE TEST SET.}
  \label{tab:Scale_comparison_online}
  \begin{tabular}{r||c||c|c|c|c|c|c}
  \Xhline{1.5pt}

  Benchmark&Method&imp surf&building&low veg&tree&car&Overall Acc.\\
  \cline{3-8}
  \Xhline{0.7pt}
  \multirow{2}*{Vaihingen}&1-scale test ({\bf `CASIA3'})&92.70&95.50&83.90&89.40&86.70&90.60 \\
  &3-scale test ({\bf `CASIA2'})&93.20&96.00&84.70&89.90&86.70&91.10 \\
  \hline
  \multirow{2}*{Potsdam}&1-scale test ({\bf `CASIA3'})&93.40&96.80&87.60&88.30&96.10&91.00 \\
  &3-scale test ({\bf `CASIA2'})&93.30&97.00&87.70&88.40&96.20&91.10 \\

  \Xhline{1.5pt}
  \end{tabular}
\end{table*}

\begin{table*}[t]
  \centering
  \scriptsize
  \caption{Ablation experiments (\%) on \emph{ISPRS Vaihingen challenge} OFFLINE VALIDATION SET. `MSC' denotes aggregating multi-scale contexts in a parallel stack shown in Fig. \ref{fig:context}(c). `MSC+SC' denotes sequentially aggregating multi-scale contexts in a self-cascaded manner. `MSC+SC+CReC' denotes sequentially aggregating multi-scale contexts in a self-cascaded manner and adding residual correction schemes in context aggregation, as shown in Fig. \ref{fig:context}(b). `Ref' denotes adding refinement. `Ref+RReC' denotes adding refinement and residual correction schemes in refinement process, as the rightmost part of Fig. \ref{fig:scasnet_overview} shows.}
  \label{tab:ablation_experiments}
  \begin{tabular}{r||cc|cc|cc|cc|cc|cc}
  \Xhline{1.5pt}

  \multirow{2}*{Model}&\multicolumn{2}{c|}{imp surf}&\multicolumn{2}{c|}{building}&\multicolumn{2}{c|}{low veg}&
  \multicolumn{2}{c|}{tree}&\multicolumn{2}{c|}{car}&\multicolumn{2}{c}{Avg.}\\
  \cline{2-13}
  &IoU&F1&IoU&F1&IoU&F1&IoU&F1&IoU&F1&mean IoU&mean F1 \\
  \Xhline{0.8pt}
  Baseline&76.74&86.84&82.33&90.31&67.77&80.79&72.62&84.14&40.71&57.86&68.04&79.99 \\
  $+$MSC&75.67&86.15&86.38&92.70&66.56&79.92&73.92&85.00&40.80&57.95&68.67&80.34 \\
  $+$MSC$+$SC&77.13&87.38&87.01&93.05&68.80&81.52&74.98&85.70&46.35&62.98&70.90&82.13 \\
  $+$MSC$+$SC$+$CReC&80.10&88.95&87.72&93.26&68.92&81.68&75.15&85.81&56.07&71.48&73.59&84.24 \\
  $+$MSC$+$SC$+$CReC$+$Ref&80.61&89.26&89.06&94.21&70.57&82.75&76.39&86.62&61.34&76.04&75.59&85.78 \\
  $+$MSC$+$SC$+$CReC$+$Ref$+$RReC&82.70&90.53&89.54&94.48&69.00&81.66&76.17&86.47&76.89&86.93&78.86&88.02 \\

  \Xhline{1.5pt}
  \end{tabular}
\end{table*}

\begin{figure*}[htp]
\centerline{\includegraphics[width=18.1cm]{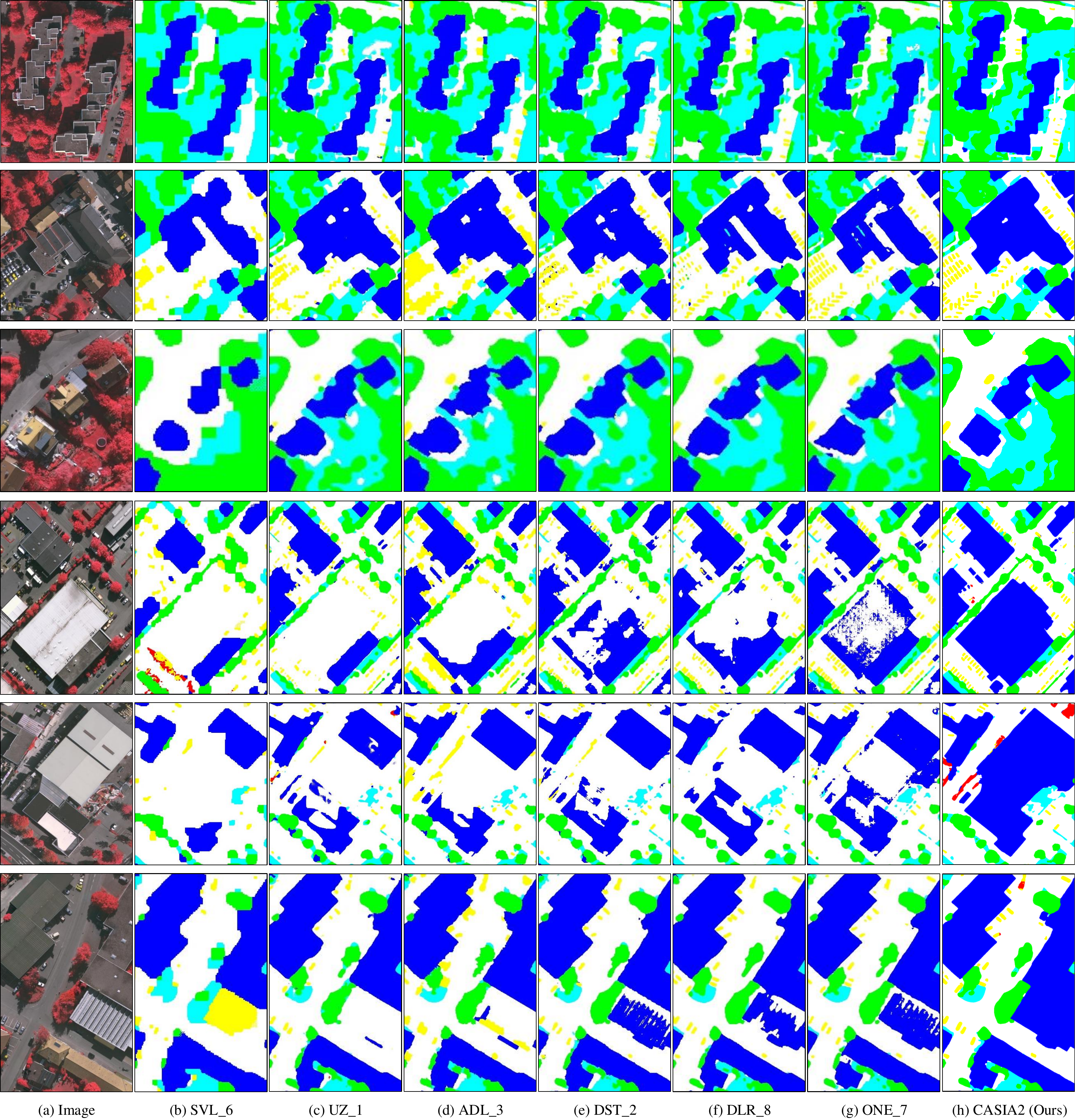}}
\caption{Qualitative comparison with other competitors' methods on \emph{ISPRS Vaihingen challenge} ONLINE TEST SET. The label includes six categories: impervious surface (imp surf, white), building (blue), low vegetation (low veg, cyan), tree (green), car (yellow) and clutter/background (red).}
\label{fig:compare_vai_online}
\end{figure*}

\begin{figure*}[htp]
\centerline{\includegraphics[width=18.1cm]{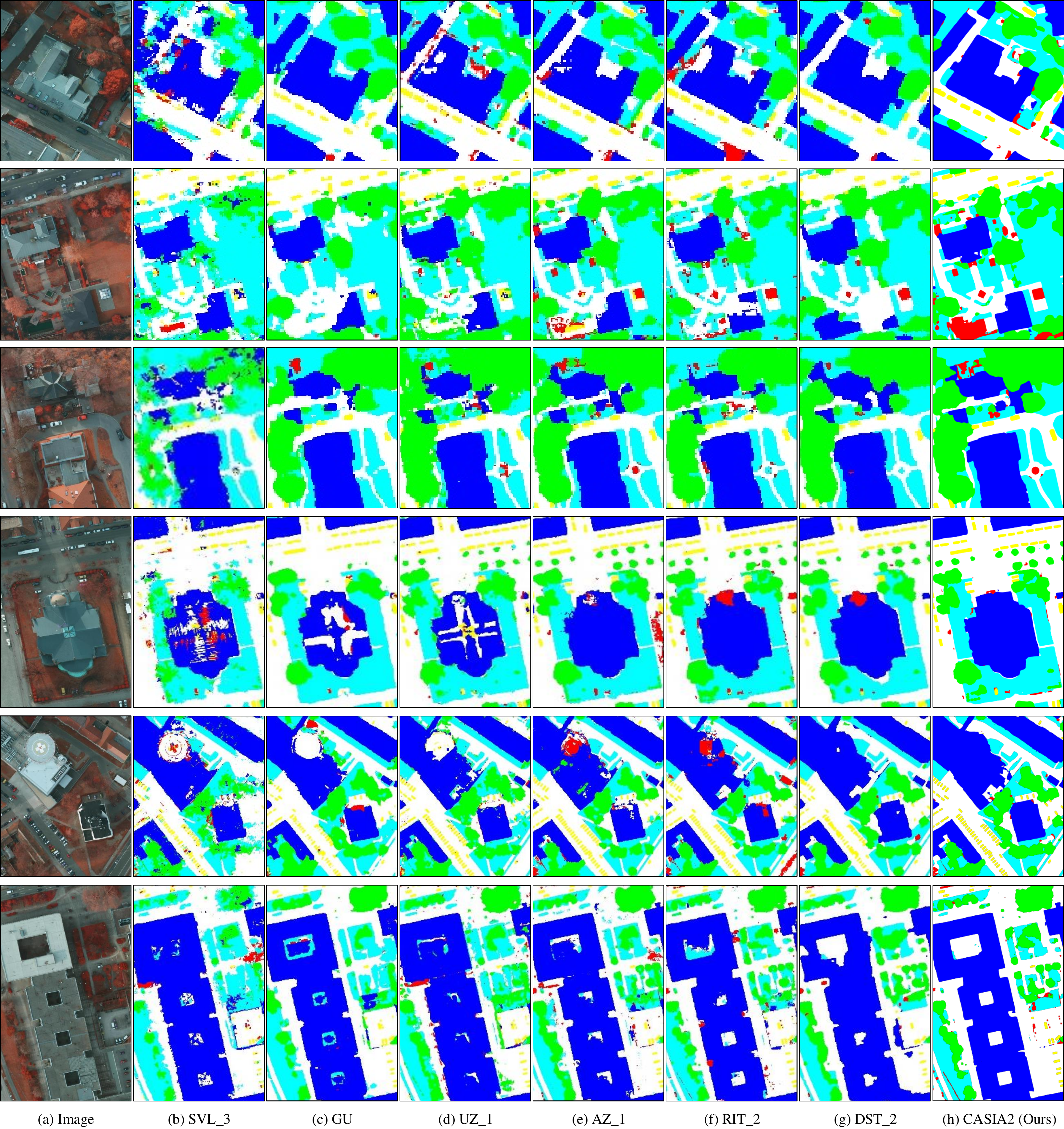}}
\caption{Qualitative comparison with other competitors' methods on \emph{ISPRS Potsdam challenge} ONLINE TEST SET. The label includes six categories: impervious surface (imp surf, white), building (blue), low vegetation (low veg, cyan), tree (green), car (yellow) and clutter/background (red).}
\label{fig:compare_pot_online}
\end{figure*}

\renewcommand{\thefootnote}{\fnsymbol{footnote}}
\setcounter{footnote}{0}
\begin{enumerate}[1)]
\item {\bf{Vaihingen Challenge}}: On benchmark test of Vaihingen\footnote{\url{http://www2.isprs.org/vaihingen-2d-semantic-labeling-contest.html}}, Fig. \ref{fig:compare_vai_online} and Table \ref{tab:Vai_comparison_online} exhibit qualitative and quantitative comparisons with different methods, respectively. As shown in Fig. \ref{fig:compare_vai_online}, other methods, even though the elevation data is used, are less effective for labeling confusing manmade objects and fine-structured objects simultaneously. In contrast, our method can obtain coherent and accurate labeling results. Moreover, our method can achieve labeling with smooth boundary and precise localization, especially for fine-structured objects like the car. As Table \ref{tab:Vai_comparison_online} shows, the quantitative performances of our method also outperform other methods by a considerable margin, especially for the car.
\setcounter{footnote}{1}
\item {\bf{Potsdam Challenge}}: On benchmark test of Potsdam\footnote{\url{http://www2.isprs.org/potsdam-2d-semantic-labeling.html}}, qualitative and quantitative comparison with different methods are exhibited in Fig. \ref{fig:compare_pot_online} and Table \ref{tab:Pot_comparison_online}, respectively. As shown in Fig. \ref{fig:compare_pot_online}, all the comparing methods obtain good results, while more coherent and accurate results are achieved by our method. In addition, our method shows better robustness to the cast shadows. Meanwhile, as can be seen in Table \ref{tab:Pot_comparison_online}, the quantitative performances of our method also outperform other methods by a considerable margin on all the categories.
\end{enumerate}

As the above comparisons demonstrate, the proposed multi-scale contexts aggregation approach is very effective for labeling confusing manmade objects. Thus, our method can perform coherent labeling even for the regions which are very hard to distinguish. Meanwhile, our refinement strategy is much effective for accurate labeling. This results in a smooth labeling with accurate localization, especially for fine-structured objects like the car. Furthermore, both of them are collaboratively integrated into a deep model with the well-designed residual correction schemes. As a result, our method outperforms other sophisticated methods by the date of submission, even though it only uses a single network based on only raw image data. Other competitors either use extra data such as DSM and model ensemble strategy, or employ structural models such as CRF.

To evaluate the performance brought by the three-scale test ( 0.5, 1 and 1.5 times the size of raw images), we submit the single scale test results to the challenge organizer. The evaluation results are listed in Table \ref{tab:Scale_comparison_online}. As can be seen, all the categories on Vaihingen dataset achieve a considerable improvement except for the car. A possible reason is that, our refinement strategy is effective enough for labeling the car with the resolution of $9$cm. Moreover, there is virtually no improvement on Potsdam dataset. Maybe for such a high resolution of $5$cm, the influence of multi-scale test is negligible.

\begin{table*}[t]
  \centering
  \scriptsize
  \caption{Quantitative comparison (\%) between with and without using finetuning technique of the encoder part on \emph{ISPRS Vaihingen challenge} OFFLINE VALIDATION SET. The model used to initialize the encoder part is pre-trained on PASCAL VOC 2012 \citep{journals_ijcv_EveringhamEGWWZ15}. `w/o' denotes without using finetuning, `w/' denotes using finetuning.}
  \label{tab:ablation_finetuning}
  \begin{tabular}{r||c||cc|cc|cc|cc|cc|cc}
  \Xhline{1.5pt}

  \multirow{2}*{Model}&\multirow{2}*{Finetuning}&\multicolumn{2}{c|}{imp surf}&\multicolumn{2}{c|}{building}&\multicolumn{2}{c|}{low veg}&
  \multicolumn{2}{c|}{tree}&\multicolumn{2}{c|}{car}&\multicolumn{2}{c}{Avg.}\\
  \cline{3-14}
  &&IoU&F1&IoU&F1&IoU&F1&IoU&F1&IoU&F1&mean IoU&mean F1 \\
  \Xhline{0.8pt}
  \multirow{2}*{VGG ScasNet}&w/o&80.55&89.23&87.23&93.18&67.18&80.06&73.94&84.88&71.35&82.88&76.05&86.05 \\
  &w/&82.70&90.53&89.54&94.48&69.00&81.66&76.17&86.47&76.89&86.93&78.86&88.02 \\
  \hline
  \multirow{2}*{ResNet ScasNet}&w/o&78.78&88.92&85.99&92.30&59.03&75.57&72.34&84.49&60.36&75.73&71.30&83.40 \\
  &w/&85.86&93.76&92.45&96.19&76.26&87.62&83.77&90.61&81.14&89.81&83.90&91.60 \\

  \Xhline{1.5pt}
  \end{tabular}
\end{table*}
\begin{table*}[t]
  \centering
  \scriptsize
  \caption{Complexity comparison (\%) with the state-of-the-art deep models.}
  \label{tab:complexity}
  \begin{tabular}{r||cccccIcc}
  \Xhline{1.5pt}

  &SegNet& FCN-8s& DeconvNet& Deeplab-ResNet& RefineNet& Ours-VGG& Ours-ResNet\\
  \hline
  Model size &112M&512M&961M&503M&234M&151M&481M \\
  Time &17s&11s&18s&47s&21s&11s&33s \\
  \Xhline{1.5pt}
  \end{tabular}
\end{table*}

\subsection{\bf Model Analysis}
\label{subsection:model_analysis}
To evaluate the performance brought by each aspect we focus on in the proposed ScasNet, the ablation experiments of VGG ScasNet are conducted. Table \ref{tab:ablation_experiments} lists the results of adding different aspects progressively. The encoder (see Fig. \ref{fig:scasnet_overview}) which is based on a VGG-Net variant \citep{conf_iclr_ChenPKMY15} is taken as the baseline. As it shows, compared with the baseline, the overall performance of fusing multi-scale contexts in the parallel stack (see Fig. \ref{fig:context}(c)) only improves slightly. By contrast, there is an improvement of near $3\%$ on mean IoU when our approach of self-cascaded fusion is adopted. Moreover, when residual correction scheme is dedicatedly employed in each position behind multi-level contexts fusion, the performance improves even more. These improvements further demonstrate the effectiveness of our multi-scale contexts aggregation approach and residual correction scheme. As can be seen, the performance of each category indeed improves when successive refinement strategy is added, but it doesn't seem to work very well. However, when residual correction scheme is elaborately applied to correct the latent fitting residual in multi-level feature fusion, the performance improves once more, especially for the car.

To evaluate the effect of \emph{transfer learning} \citep{conf_nips_JasonJYH2014,conf_cvpr_PenaANKSJ2015,journals_rs_FanHu15,journals_corr_XieNMLDES2015}, which is used for training ScasNet, the quantitative performance brought by initializing the encoder's parameters (see Fig. \ref{fig:scasnet_overview}) with pre-trained model (i.e., finetuning) are listed in Table \ref{tab:ablation_finetuning}. As it shows, the performance of VGG ScasNet improves slightly, while ResNet ScasNet improves significantly. These results indicate that, it is very difficult to train deep models sufficiently with so small remote sensing datasets, especially for the very deep models, e.g., the model based on $101$-layer ResNet. Therefore, the ScasNet benefits from the widely used \emph{transfer learning} in the field of \emph{deep learning}.
\begin{figure*}[t]
\begin{minipage}[b]{1\linewidth}
  \centering
  \centerline{\includegraphics[width=14.0cm]{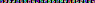}}
  \vspace{-5pt}
  \centerline{(a)}
\end{minipage}
\begin{minipage}[b]{1\linewidth}
  \centering
  \centerline{\includegraphics[width=14.0cm]{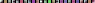}}
  \vspace{-5pt}
  \centerline{(b)}
\end{minipage}
\begin{minipage}[b]{1\linewidth}
  \centering
  \centerline{\includegraphics[width=14.0cm]{Figures/feature_map.pdf}}
\end{minipage}

\caption{Filters and feature maps learned by VGG ScasNet. For better visualization, the visuals are mapped to full channel range and combined in some cases to occupy all RGB channels. In these features, colored regions denote strong responses and deep black regions otherwise. (a) The 1st-layer convolutional filters ($3\times3\times3\times64$) without finetuning. (b) The 1st-layer convolutional filters ($3\times3\times3\times64$) with finetuning. (c) The last convolutional feature maps in the 1st stage ($400\times400\times64$). (d) The last convolutional feature maps in the 2nd stage ($201\times201\times128$). (e) Feature maps outputted by the encoder ($51\times51\times512$). (f) Feature maps after our multi-scale contexts aggregation approach ($51\times51\times512$). (g) Feature maps after our refinement strategy ($101\times101\times256$). (h) Fused feature maps before residual correction ($101\times101\times256$). (i) Feature maps learned by inverse residual mapping $\mathcal{H}[\cdot]$ (see Fig. \ref{fig:residual}). (j) Fused feature maps after residual correction ($101\times101\times256$).}
\label{fig:featuremap}
\end{figure*}

To further verify the validity of each aspect of our ScasNet, features of some key layers in VGG ScasNet are visualized in Fig. \ref{fig:featuremap}. For clarity, we only visualize part of features in the last layers before the pooling layers, more detailed visualization can be referred in the Appendix B of supplementary material. As shown in Fig. \ref{fig:featuremap}(a) and (b), the 1st-layer convolutional filters tend to learn more meaningful features after funetuning, which indicates the validity of \emph{transfer learning}. As Fig. \ref{fig:featuremap}(c) and (d) indicate, the layers of the first two stages tend to contain a lot of noise (e.g., too much littery texture), which could weaken the robustness of ScasNet. That is a reason why they are not incorporated into the refinement process.

As can be seen in Fig. \ref{fig:featuremap}(e), the responses of feature maps outputted by the encoder tend to be quite messy and coarse. However, as shown in Fig. \ref{fig:featuremap}(f), coherent and intact semantic responses can be obtained when our multi-scale contexts aggregation approach is used. Moreover, as Fig. \ref{fig:featuremap}(g) shows, much low-level details are recovered when our refinement strategy is used. The boundary responses of cars and trees can be clearly seen.

Fig. \ref{fig:featuremap}(h), (i) and (j) visualize the fused feature maps before residual correction, the feature maps learned by inverse residual mapping $\mathcal{H}[\cdot]$ (see Fig. \ref{fig:residual}) and the fused feature maps after residual correction, respectively. As Fig. \ref{fig:featuremap}(h) shows, there is much information lost when two feature maps with semantics of different levels are fused. Nevertheless, as shown in Fig. \ref{fig:featuremap}(j), these deficiencies are mitigated significantly when our residual correction scheme is employed. That is, as Fig. \ref{fig:featuremap}(i) shows, the inverse residual mapping $\mathcal{H}[\cdot]$ could compensate for the lack of information, thus counteracting the adverse effect of the latent fitting residual in multi-level feature fusion.

Table \ref{tab:complexity} compares the complexity of ScasNet with the state-of-the-art deep models. The time complexity is obtained by averaging the time to perform single scale test on $5$ images (average size of $2392\times2191$ pixels) with a GTX Titan X GPU. As it shows, ScasNet produces competitive results on both space and time complexity.

\section{Conclusion}
\label{Section5}
In this work, a novel end-to-end self-cascaded convolutional neural network (ScasNet) has been proposed to perform semantic labeling in VHR images. The proposed ScasNet achieves excellent performance by focusing on three key aspects: 1) A self-cascaded architecture is proposed to sequentially aggregate global-to-local contexts, which are very effective for confusing manmade objects recognition. Technically, multi-scale contexts are first captured on the output of a CNN encoder, and then they are successively aggregated in a self-cascaded manner; 2) With the acquired contextual information, a coarse-to-fine refinement strategy is proposed to progressively refine the target objects using the low-level features learned by CNN's shallow layers. Therefore, the coarse labeling map is gradually refined, especially for intricate fine-structured objects; 3) A residual correction scheme is proposed for multi-feature fusion inside ScasNet. It greatly corrects the latent fitting residual caused by the semantic gaps in features of different levels, thus further improves the performance of ScasNet. As a result of these specific designs, ScasNet can perform semantic labeling effectively in a manner of global-to-local and coarse-to-fine.

Extensive experiments verify the advantages of ScasNet: 1) On both quantitative and visual performances, ScasNet achieves extraordinarily more coherent, complete and accurate labeling results while remaining better robustness to the occlusions and cast shadows than all the comparing advanced deep models; 2) ScasNet outperforms the state-of-the-art methods on two challenging benchmarks by the date of submission: \emph{ISPRS 2D Semantic Labeling Challenge} for Vaihingen and Potsdam, even not using the available elevation data, model ensemble strategy or any postprocessing; 3) ScasNet also shows extra advantages on both space and time complexity compared with some complex deep models.
\\
\\
\textbf{Acknowledgments}: The authors wish to thank the editors and anonymous reviewers for their valuable comments which greatly improved the paper's quality. The authors also wish to thank the ISPRS for providing the research community with the awesome challenge datasets, and thank Markus Gerke for the support of submissions. This work was supported by the National Natural Science Foundation of China under Grants 91646207, 61403375, 61573352, 61403376 and 91438105.

\newpage
\setlength{\bibspacing}{\baselineskip}
{\footnotesize{}\bibliographystyle{elsarticle-harv}
\bibliography{ref}
}{\footnotesize \par}


\end{document}